\documentclass{elsart}               

\usepackage{natbib}

\usepackage{amsmath}
\usepackage{mathtools}
\usepackage{amssymb} 
\usepackage{upgreek}
\usepackage{natbib}
\usepackage{xcolor}
\usepackage{comment}
\usepackage{wrapfig}
\usepackage{comment}

\usepackage{graphicx}          

\theoremstyle{definition}
\newtheorem{definition}{Definition}[section]              

\theoremstyle{assumption}
\newtheorem{assumption}{Assumption}[section]

\begin{document}

\begin{frontmatter}

\title{Geometric Control of Mechanical Systems with Symmetries Based on Sliding Modes\thanksref{footnoteinfo}} 

\thanks[footnoteinfo]{This paper was not presented at any IFAC 
meeting. Corresponding author Yu~Tang.}

\author[ZJU,NIT]{Eduardo Espindola}\ead{eespindola@zju.edu.cn},    
\author[NIT]{Yu Tang}\ead{tangy@nbt.edu.cn},               

\address[ZJU]{College of Electrical Engineering, Zhejiang University, Hangzhou, China} 
\address[NIT]{School of Information Science and Engineering, NingboTech University, Ningbo, China}


\begin{abstract}                          
In this paper, we propose a framework for designing sliding mode controllers for a class of mechanical systems with symmetry, both unconstrained and constrained, that evolve on principal fiber bundles. Control laws are developed based on the reduced motion equations by exploring symmetries, leading to a sliding mode control strategy where the reaching stage is executed on the base space, and the sliding stage is performed on the structure group. Thus, design complexity is reduced, and difficult choices for coordinate representations when working with a particular Lie group are avoided. For this purpose, a sliding subgroup is constructed on the structure group based on a kinematic controller, and the sliding variable will converge to the identity of the state manifold upon reaching the sliding subgroup. A reaching law based on a general sliding vector field is then designed on the base space using the local form of the mechanical connection to drive the sliding variable to the sliding subgroup, and its time evolution is given according to the appropriate covariant derivative. Almost global asymptotic stability and local exponential stability are demonstrated using a Lyapunov analysis. We apply the results to a fully actuated system (a rigid spacecraft actuated by reaction wheels) and a subactuated nonholonomic system (unicycle mobile robot actuated by wheels), which is also simulated for illustration.  
\end{abstract}

\end{frontmatter}

\section{Introduction}
For robotic systems that evolve on a general manifold considered in this paper, many existing control strategies have been developed in the Euclidean configuration space, that is, a planar configuration space with zero curvature \citep{hassan2002nonlinear}, 
either by embedding the configuration space into a higher-dimensional Euclidean space or choosing a local coordinate chart of the configuration manifold as the configuration space. The former can lead to the loss of structural properties of a manifold, such as symmetry, which would be otherwise exploited for design purposes \citep{bloch1996nonholonomic,bonnabel2008symmetry}; the latter immediately sacrifices the notion of globality and consequent robustness. In addition, for a given smooth manifold, it is unclear how to choose a local representation, as there are multiple options for configuration spaces with varying dimensions (\citep{bullo2019geometric}).

To address these challenges, the geometric approach, which focuses on coordinate-free designs, has been drawing much attention from the control community for controller and observer development. Using tools from differential geometry, controllers (\cite{bullo1999tracking,maithripala2006almost}) and observers (\cite{aghannan2003intrinsic,bonnabel2008symmetry}) were developed on generic configuration manifolds, allowing one to take advantage of important geometric features, such as symmetries and invariance.  The work of \cite{bullo1999tracking,bloch1995nonholonomic,hussein2008optimal} developed control designs on a general Riemannian manifold. Using results from the general Riemannian manifold specified to Lie groups, \cite{de2018output,maithripala2015intrinsic,saccon2013optimal,bullo2000controllability,basso2023synergistic} developed tracking controllers by stabilizing the error dynamics defined naturally on the Lie group. 

Sliding mode control is a widely known control strategy for its salient features, such as robustness to uncertainties and disturbances that satisfy a matching condition \citep{utkin1977variable}. Starting from an initial condition, the controller drives the system trajectories to the sliding surface designed to meet some specific criteria in the reaching stage. Upon reaching the sliding surface, the constrained trajectories converge to the origin, achieving the control objective. 
This control strategy, originally developed for systems evolving in Euclidean space, has been recently extended to a general Lie group in \cite{espindola2023geometric}, where a sliding subgroup is properly defined globally on the state manifold, i.e., the isomorphism of the tangent bundle of the underlying Lie group. Therefore, almost global asymptotic stability and local exponential stability can be achieved.  

However, the configuration of a robotic system is often a larger manifold than just a Lie group and motions may be constrained. Taking the unicycle as an example, its configuration consists of a Lie group ($SE(2)$), which represents the pose, and a manifold (torus), which describes the wheel angles. Also, nonholonomic constraints like no-slip conditions make the velocity space a distribution rather than the entire tangent bundle. This has raised the interest of considering mechanical systems that evolve on a {\it trivial principal fiber bundle}, i.e., a configuration manifold that can be naturally split into the pose group (a Lie group) and a base manifold (a Riemannian manifold). In addition, the bundle structure helps incorporate these constraints naturally through the mechanical connection.
In practice, many configurations of robotic systems fall into this category, such as snake-shaped robots \citep{mcisaac2003motion,shammas2007geometric,guo2018guided}, wheeled robots \citep{martins2008adaptive,hwang2016comparison}, legged robots \citep{spong2005controlled,reher2021dynamic}, finned robots \citep{morgansen2007geometric,wu2017development}, among others. The work of \cite{koon1997optimal,bloch1996nonholonomic,ostrowski1999computing,cortes2002simple} presented a framework based on Lagrangian reduction for analysis and design on a principal fiber bundle that allows achieving the desired configuration by manipulating the shape variables.  

Based on this framework, we develop in this paper a geometric control design based on sliding modes for such systems, both unconstrained and under nonholonomic constraints. By exploring symmetries, control laws are developed based on the reduced motion equations, leading to a sliding control strategy where the reaching stage is executed on the base space and the sliding stage is performed on the structure group. In this way, design complexity is reduced and, more importantly, difficult choices for coordinate representations when working with a particular Lie group are avoided. Furthermore, for systems evolving on a Lie group with a non-symmetric potential energy, it is possible to rearrange the configuration manifold to have a principal bundle structure, in which the structure group has a symmetric Lagrangian in a reduced-dimension Lie group, so that the present design framework can be applied \citep{bloch1996nonholonomic}.

The main contributions are summarized as follows.
\begin{enumerate}
    \item A framework for a systematic design of a sliding mode controller for unconstrained mechanical systems on a trivial principal fiber bundle. A sliding vector field on the base (shape) space is introduced, which enables us to develop a reaching law on the base space to drive, through the local form of the mechanical  connection, the sliding variable on the pose group to the sliding subgroup. The intrinsic evolution of the sliding vector field is calculated by using an appropriate affine connection for the reduced system. Upon reaching the sliding subgroup, the sliding variable converges to the identity of the state manifold, achieving the control objective.
    \item  This systematic design is further extended for mechanical systems on a trivial principal fiber bundle with (symmetric) nonholonomic constraints by considering a nonholonomic affine connection on the constraint distribution. A novel auxiliary variable is introduced to bypass the problem that velocity tracking error generally does not belong to the constrained distribution due to the adjoint operation. 
    \item The application of the proposed design framework to a fully actuated (a rigid spacecraft actuated by reaction wheels) and 
    a nonholonomic (an unicycle mobile robot) system is developed, resulting in a novel sliding mode controller. Numerical simulations are included for illustration and comparisons. 
\end{enumerate}

The remainder of the paper is organized as follows. In Section \ref{sec:Model}, brief mathematical preliminaries are given, and affine connections and the reduced dynamics of mechanical systems with symmetries are described. Section \ref{sec:UnCtrl} is devoted to the design of controllers for unconstrained systems, while Section \ref{sec:CnCtrl} extends the previous controller design to constrained systems. The theoretical results were applied to two systems in Section \ref{sec:Examp}, where numerical simulations are presented for illustration. 
Finally, conclusions are given in Section \ref{sec:Concl}. In the Appendix, some technical results are described.

\subsection{Notation}

Following the notation in \cite{abraham2008foundations,bullo2019geometric}, the symbol $\langle\langle \cdot, \cdot\rangle\rangle \in \mathbb{R}$ is used for an inner product in the tangent space of a smooth manifold $\mathcal{M}$. A set of intervals is denoted by $\mathbf{I}$. For a $\mathbb{R}$-vector space $V$,  $V^{*}$ denotes its dual space. If $B\vcentcolon V\times V \to \mathbb{R}$ is a bilinear map, the flat map $B^{\flat}\vcentcolon V\to V^{*}$ is defined as $ \langle B^{\flat}(v);u \rangle =B(u,v)$, $\forall u,v \in V$, $B^{\flat}(v)\in V^{*}$, where $\langle \alpha ; u \rangle = \alpha(u)$ denotes the image in $\mathbb{R}$ of $u\in V$ under the covector $\alpha \in V^{*}$. The inverse map, known as the sharp map, is denoted by $B^{\sharp}\vcentcolon V^{*} \to V$. Given a smooth manifold $Q$, $TQ$ denotes the tangent bundle $Q\times T_{q}Q$, $\forall q\in Q$, with $T_{q}Q$ being the tangent space in $q\in Q$.  The set of smooth functions and vector fields on $Q$ is denoted by $\mathcal{C}^{\infty}(Q)$ and $\Gamma^{\infty}(TQ)$, respectively. The assignment $\nabla\vcentcolon \Gamma^{\infty}(TQ)\times\Gamma^{\infty}(TQ)\to \Gamma^{\infty}(TQ)$ describes a $\mathbb{R}$-bilinear affine connection. For any pair $X,Y \in \Gamma^{\infty}(TQ)$, $[X,Y]\in\Gamma^{\infty}(TQ)$ denotes the Jacobi-Lie bracket, and equivalently, the Lie derivative $\mathcal{L}_{X}Y \in \Gamma^{\infty}(TQ)$ of $Y$ with respect to $X$. For a Lie group $G$, $e\in G$ is the identity of the group, and $\mathfrak{g}$ is the tangent space in $e$, i.e., the Lie algebra. The adjoint operator $\mathrm{ad}\vcentcolon \mathfrak{g}\times \mathfrak{g}\to \mathfrak{g}$ represents the operation $\mathrm{ad}_{\zeta}\xi = [ \zeta, \xi ]$, while its dual $\mathrm{ad}^{*}_{\zeta}\vcentcolon \mathfrak{g}^{*} \to \mathfrak{g}^{*}$ is given by $\langle \mathrm{ad}^{*}_{\zeta}\alpha ; \xi \rangle = \langle \alpha ; [\zeta ,\xi] \rangle$, for all $\zeta,\xi\in\mathfrak{g}$, $\alpha\in\mathfrak{g}^{*}$. In addition, the adjoint action $\mathrm{Ad}\vcentcolon G\times\mathfrak{g} \to \mathfrak{g}$ describes the automorphism of the Lie algebra $\mathrm{Ad}_{g}\vcentcolon \mathfrak{g}\to\mathfrak{g}$ according to $\mathrm{Ad}_{g}\xi = g\xi g^{-1}$, $\forall g\in G$. Given a basis $\{\underline{e}_{a}\}^{k}_{a=1}$ of $\mathfrak{g}$, the structure constants are denoted by $C^{a}_{bc}$, such that $[\underline{e}_{b},\underline{e}_{c}] = C^{a}_{bc}\underline{e}_{a}$.

\section{Mechanical systems on principal fiber bundles}\label{sec:Model}
\subsection{Motion equations}
A simple mechanical system is represented by a tuple $\{ Q,\mathbb{G},V,\mathcal{F}\}$, where $Q$ is the configuration manifold of dimension $n$, $\mathbb{G}$ is a Riemannian metric, $V\in\mathcal{C}^{\infty}(Q)$ is a potential function, 
and $\mathcal{F}=\{ f^{1},\ldots , f^{m}  \}$, with $m\leq n$, is the control force. Thus, given a differentiable curve $\gamma\vcentcolon \mathbf{I}\to Q$, with $\dot{\gamma}(t)\in T_{\gamma(t)}Q$ a vector field in the tangent space $T_{\gamma(t)}Q$ on the curve $\gamma (t)\in Q$, $\forall t\in\mathbf{I}$, the intrinsic equations of motion for a simple mechanical are described by the Levi-Civita connection $\overset{\mathbb{G}}{\nabla}$, which is an affine connection, i.e., a connection which is both torsion free and compatible with the metric $\mathbb{G}$ \citep{bullo2019geometric}
\begin{equation}\label{eq:UnMechSys}
    \overset{\mathbb{G}}{\nabla}_{\dot{\gamma}(t)}\dot{\gamma}(t) = -\mathrm{grad}V( \gamma) + \sum^{m}_{a=1}u^{a}(t)F_{a}\left(\gamma (t)\right) ,
\end{equation}
where $\mathrm{grad}V\left( \gamma\right) = \mathbb{G}^{\sharp}\left( \mathrm{d}V(\gamma)\right)$ is the gradient of $V$, $u\vcentcolon \mathbf{I}\to U\subset \mathbb{R}^m$ are locally integrable controls, and $F_{a}(\gamma) = \mathbb{G}^{\sharp}\left( f^{a}\left( \gamma\right) \right) \in\Gamma^{\infty}(TQ)$ are the vector-field representation of the control forces $f^{a}\in\Gamma^{\infty}(T^{*}Q)$. 

Likewise, a {\it constrained} mechanical system is represented by a tuple $\{ Q,\mathbb{G},V,\mathcal{F},\mathcal{D} \}$, being $\mathcal{D}\subset TQ$ the (nonholonomic) constraint distribution on the tangent bundle $TQ$ of dimension $(n-l)$, which can be locally described by $l$ independent functions $\omega_{j}(q)\dot{q}$, $1\leq j \leq l$. Let $\mathcal{P}\vcentcolon TQ \to \mathcal{D}$ and $\mathcal{Q}:TQ\to \mathcal{D}^{\perp}$ be 
the complementary $\mathbb{G}$-orthogonal projections of $\mathcal{D}$, then the motion equations for a constrained system are given by
\begin{equation}\label{eq:CnMechSys}
    \overset{\mathbb{G}}{\overline{\nabla}}_{\dot{\gamma}(t)}\dot{\gamma}(t) = -\mathcal{P}\left( \mathrm{grad}V( \gamma) \right) + \sum^{m}_{a=1}u^{a}(t)\mathcal{P}\left( F_{a}\left(\gamma\right)\right) ,
\end{equation}
with $\dot\gamma(0)\in \mathcal{D}_{\gamma(0)} $, which ensures that  $\dot{\gamma}(t)\in\mathcal{D}_{\gamma (t)}$, $\forall t\geq 0$,  where $\overset{\mathbb{G}}{\overline{\nabla}}\vcentcolon \Gamma^{\infty}(TQ)\times \mathcal{D} \to \mathcal{D}$ is the nonholonomic affine connection -- the connection $ \overset{\mathbb{G}}{\nabla}$ restricts to  the distribution $\mathcal{D}$ \citep{bloch1996nonholonomic}. 

\subsection{Principal fiber bundles}

A principal fiber bundle is defined by a tuple $\left(Q,G,\Phi,M\right)$, where the total space $Q$ and the base space $M$ are smooth manifolds, $G$ is a Lie group, and $\Phi$ is a free, proper and smooth left (right) action of $G$ on $Q$. 

For easy exposition, we assume that the configuration manifold of simple mechanical systems is the total space of  {\it trivial} principal fiber bundle, i.e., it admits the natural splitting $Q=G\times M$, where $G$ is the space for pose coordinates (fiber group), and $M$ corresponds to the shape space (base space). For systems on a general principal bundle, one can always trivialize $Q$ and work locally with the resulting trivialized principal bundle (\cite{cortes2002simple}).  Then each coordinate $q=(g,r)\in Q$ is represented by the pose coordinates $g\in G$ and the shape coordinates $r\in M$. Furthermore, according to the left group action%
$ \left( g,q\right) \in G\times Q \mapsto \Phi\left( g,q\right) = \Phi_{g}(q) = gq \in Q$, the orbit through a point $q_{0}\in Q$ is $\mathcal{O}_{G}(q_{0})=\{ gq_{0}\;|\; g\in G \}$. The tangent space at $q\in Q$ is given by $T_{q}\mathcal{O}_{G}(q)=\{ \xi_{Q}\left(q\right)\;|\; \xi\in \mathfrak{g} \}$ being $\xi_{Q}\in \Gamma^{\infty}(TQ)$ the infinitesimal generator for the corresponding action $\Phi$. Then a {\it vertical} distribution $\mathcal{V}$ on $Q$ can be defined as the set of tangent vectors to the orbits of $G$ on $Q$, i.e., $\mathcal{V}_{q}=T_{q}\mathcal{O}_{G}(q)$, $\forall q\in Q$. Likewise, the {\it horizontal} distribution $\mathcal{H}$ is the $\mathbb{G}$-orthogonal complement of $\mathcal{V}$, such that $TQ = \mathcal{V}\oplus \mathcal{H}$, namely, for any $X_{q}\in T_{q}Q$, it holds $X_{q} = X_{q}^{\mathrm{h}} + X_{q}^{\mathrm{v}}$, with $X_{q}^{\mathrm{h}}\in\mathcal{H}_{q}$ and $X_{q}^{\mathrm{v}}\in \mathcal{V}_{q}$.

In addition, the horizontal distribution $\mathcal{H}$ can be described by the principal connection of the fiber bundle as $\mathcal{H}_{q} =\{ Y_{q}\in T_{q}Q \; |\; \mathcal{A}(Y_{q})=0 \}$, where $\mathcal{A}\vcentcolon TQ \to \mathfrak{g}$ is the {\it principal connection}, which is equivariant in the sense that $\mathcal{A}\left(\left(\Phi_{g}\right)_{*}X\right) = \mathrm{Ad}_{g}\mathcal{A}(X)$, $\forall X\in TQ$, and satisfies $\mathcal{A}\left( \xi_{Q}(q) \right) = \xi$, $\forall \xi\in\mathfrak{g}$. Consequently, for $q=(g,r)\in Q$ and $\dot{q} = \left( \dot{g},\dot{r}\right) \in T_{q}Q$, it follows from \cite{bloch1996nonholonomic} that
\begin{equation}\label{eq:Aconn}
    \mathcal{A}\left(q,\dot{q}\right)  = \mathcal{A}\left(g,r,\dot{g}, \dot{r}\right) = \mathrm{Ad}_{g}\left( \xi + A(r)\dot{r} \right),
\end{equation}
where $\xi = g^{-1}\dot{g}$, and $A(r)\vcentcolon T_{r}M\to \mathfrak {g}$ is the {\it local form} of the principal connection $\mathcal{A}$. 

Given a distribution $\mathcal{D}\subset TQ$ such that $\mathcal{D}+\mathcal{V}=TQ$, a section $\overline{\mathcal{V}}_{q} = \mathcal{V}_{q}\cap \mathcal{D}_{q}$ 
and $\overline{\mathcal{H}}_{q} = \overline{\mathcal{V}}^{\perp}_{q}\cap \mathcal{D}_{q}$  can be considered \citep{cortes2002simple}, where $\overline{\mathcal{V}}^{\perp}_{q}$ is the $\mathbb{G}$-orthogonal complement of $\overline{\mathcal{V}}_{q}$ for each $q\in Q$. Likewise, the set of elements of the Lie algebra restricted to the distribution $\overline{\mathfrak{g}}$ is determined by $\overline{\mathfrak{g}}_{q} = \{ \xi\in\mathfrak{g} \;|\; \xi_{Q}(q)\in \overline{\mathcal{V}}_{q} \}$. 
Let $\overline{\mathcal{A}}\vcentcolon TQ\to \overline{\mathfrak{g}}$ denote the constrained principal connection to describe $\overline{\mathcal{H}}$, then 
$\overline{\mathcal{A}}=\mathcal{A}^{\mathrm{sym}}+\mathcal{A}^{\mathrm{skw}}$, where $\mathcal{A}^{\mathrm{sym}}\vcentcolon T_{q}Q \to \overline{\mathcal{V}}_{q}$, and $\mathcal{A}^{\mathrm{skw}}\vcentcolon T_{q}Q \to \overline{\mathcal{V}}^{\perp}_{q}$.

\subsection{Lagrangian reduction}

Let $\Pi=\left( Q, G,\Phi, M\right)$ be a trivial principal fiber bundle, then the system $\Sigma=\{Q,\mathbb{G},V,\mathcal{F}\}$ is invariant under the action $\Phi$ if $\Phi^{*}_{g}\mathbb{G} = \mathbb{G}$, $V\circ \Phi_{g} = V$, and $\Phi^{*}_{g}f^{a} = f^{a}$, for all $g\in G$, and $1\leq a\leq m$. $\Phi$ is termed a {\it symmetry} of the mechanical system $\Sigma$. Likewise, a constrained mechanical system $\Sigma_{\mathcal{D}}=\{Q,\mathbb{G},V,\mathcal{F},\mathcal{D}\}$ is invariant under $\Phi$ if, in addition, $T_{q}\Phi_{g}\left(\mathcal{D}_{q}\right) = \mathcal{D}_{\Phi_{g}(q)}$, for all $q\in Q$.

The symmetry of a mechanical system allows us to work with (simplified) reduced equations that are useful for analysis and controller designs.
The reduced equations can be obtained directly from the {\it reduced Lagrangian} \citep{ostrowski1999computing}
\begin{equation}\label{eq:lag}
 l \left( r,\xi,\dot{r}\right) = \frac{1}{2}\mathcal{G}\left( \xi,\dot{r},\xi,\dot{r} \right) - V(r)   
\end{equation}
of the Lagrangian $L\left(q,\dot{q}\right) = \frac{1}{2}\mathbb{G}(q)\left( \dot{q},\dot{q}\right) - V(q)$ by writing $ L\left(q,\dot{q}\right) = L\left( g,r,\dot g,\dot{r}\right) = L\left( e,r,\xi,\dot{r}\right) = l \left( r,\xi,\dot{r}\right)$ with the {\it reduced metric} $\mathcal{G}$, which can be expressed in matrix form as 
\begin{equation}\label{eq:rMetric}
    [\mathcal{G}] = \left[ \begin{array}{cc}
       I(r) & I(r)A(r) \\
        A^{T}(r)I(r) &  m(r)
    \end{array} \right],
\end{equation}
where $I(r)$ and $A(r)$ are the local expressions of the locked inertia tensor $\mathcal{I}(q)$ 
and the mechanical connection $\mathcal{A}^{\mathrm{m}}(\dot q)$, respectively, which depend only on the base variable $r$, and $m(r)$ represents the system inertia for a fixed shape $r\in M$. The locked inertia tensor $\mathcal{I}(q)\vcentcolon \mathfrak{g}\to\mathfrak{g}^{*}$ is defined by $\langle \mathcal{I}(q)\xi ;\zeta \rangle = \mathcal{G}\left( \xi_{Q}(q),\zeta_{Q}(q)\right)$, and calculated as $\mathcal{I}(g,r)=\mathrm{Ad}^{*}_{g^{-1}}I(r)\mathrm{Ad}_{g^{-1}}$, $\forall q=(g,r)\in Q$. 
The local form $A(r)$ determines how changes in internal shape create motions of the net system, and is defined through the mechanical connection $\mathcal{A}^{\mathrm{m}}(\dot{q})\triangleq\mathcal{I}^{-1}(q)J(\dot{q})$,  $\forall \dot{q}\in T_{q}Q$, where $J\vcentcolon TQ\to \mathfrak{g}^{*}$ is the momentum map $\langle J(\dot{q});\xi \rangle = \langle \frac{\partial L}{\partial \dot{q}}\left( \dot{q} \right) ;  \xi_{Q}(q) \rangle$. Consequently, $\mathcal{A}^{\mathrm{m}}(\dot{q})$ is a principal connection for the fiber bundle $\Pi$ and the system $\Sigma$. In a body-fixed coordinate frame, it allows defining the generalized momentum $p\in \mathfrak{g}^{*}$, according to \eqref{eq:Aconn} and \eqref{eq:lag}, as
\begin{equation}\label{eq:p}
    p \triangleq \frac{\partial l}{\partial \xi} = I(r)\xi + I(r)A(r)\dot{r}.
\end{equation}

Note that by considering the locked body velocity $\Omega = \xi + A(r)\dot{r}$ in the reduced Lagrangian $l\left( r,\Omega,\dot{r}\right) $ in Eq. \eqref{eq:lag}, the reduced metric \eqref{eq:rMetric} admits a block diagonalization
\begin{equation}\label{eq:rDMetric}
    \left[\hat{\mathcal{G}}\right] = 
       \begin{bmatrix}I(r) & 0 \\
        0 & \mathbb{M}(r) 
    \end{bmatrix},
\end{equation}
with {\it reduced mass matrix} $\mathbb{M}(r)=m(r) - A(r)^{T}I(r)A(r)$. 

For a symmetric constrained system $\Sigma_{\mathcal{D}}=\{Q,\mathbb{G},V,\mathcal{F},\mathcal{D}\}$ a locked inertia tensor $\overline{\mathcal{I}}\vcentcolon \overline{\mathfrak{g}}\to \overline{\mathfrak{g}}^{*}$ and a constrained momentum map $\overline{J}\vcentcolon TQ\to \overline{\mathfrak{g}}^{*}$ are considered, which fulfill the following 
\begin{equation*}
 \mathcal{A}^{\mathrm{sym}}(\dot{q})\triangleq \left(\overline{\mathcal{I}}^{-1}(q)\overline{J}(q,\dot{q})\right)_{Q}. 
\end{equation*}
In addition, $\mathcal{A}^{\mathrm{skw}}(q)\dot{q}= 0$ describes the constraints in a kinematic relation ($\omega_{j}(q)\dot{q}=0$). Thus, the nonholonomic connection for the system $\Sigma_{\mathcal{D}}$ is $\mathcal{A}^{\mathrm{nh}}=\mathcal{A}^{\mathrm{sym}}+\mathcal{A}^{\mathrm{skw}}$. Moreover, assume $\overline{\mathfrak{g}}_{(e,r)}=\mathrm{span}\{ \underline{e}_{1},\underline{e}_{2},\ldots , \underline{e}_{n-l} \}$, being $\{\underline{e}_{i}\}^{n-l}_{i=1}$ a $\mathbb{G}$-orthogonal basis restricted to $\mathcal{V}$, then the generalized momentum $\bar{p}\in\overline{\mathfrak{g}}^{*}$ for the constrained system is 
\begin{equation}\label{eq:pbdef}
    \bar{p}_{i} \triangleq \left\langle \frac{\partial l}{\partial \xi} ; \underline{e}_{i}(r) \right\rangle ,\quad 1\leq i\leq n-l.
\end{equation}
Consequently, 
\begin{equation}\label{eq:pb}
    \bar{p} = \bar{I}(r)\xi + \bar{I}(r)\mathbb{A}(r)\dot{r},
\end{equation}
where $\bar{I}(r)$ and $\mathbb{A}(r)$ are the local expressions of $\overline{\mathcal{I}}(\gamma)$ and $\mathcal{A}^{\mathrm{nh}}$, respectively.

\subsection{Levi-Civita connection under symmetry}

The Levi-Civita connection and the nonholonomic affine connection can be decomposed according to the principal fiber bundle structure of the total space $Q$ (cf. Appendix \ref{App:LCterms}). 
Therefore, given a principal fiber bundle $\Pi=\big( Q, G,\Phi,$ $ M\big)$, an invariant mechanical system $\Sigma=\{Q,\mathbb{G},V,\mathcal{F}\}$, a curve 
$\gamma (t) = (g(t),r(t))\in Q$, and its time change rate $\dot{\gamma}(t)=(g\xi(t) , \dot{r}(t))\in T_{\gamma (t)}Q$, the reduced motion equations can be written as  
\begin{align}
    \overset{\mathbb{M}}{\nabla}_{\dot{r}(t)}\dot{r}(t) &= \frac{1}{2}\mathbb{M}^{\sharp}(h(\dot{\gamma}(t),\dot{\gamma}(t))) -\mathrm{grad}V(r) + f_{u} ,\label{eq:ShapeDyn} \\ 
    \dot{p}(t) &= \mathrm{ad}^{*}_{\xi}p(t) + \tau_{G}, \label{eq:pp} \\
    \mathfrak{g}\ni \xi (t) &\triangleq g^{-1}\dot{g}(t) =  - A(r)\dot{r}(t) + I^{-1}(r) p(t), \label{eq:xi}  
\end{align}
where $f_{u}=\sum^{m}_{a=1} u^{a}(t)\mathbb{M}^{\sharp}\left( \tau^{a}_{M} - \tau^{j}_{G}A^{a}_{j}\right) \in \Gamma^{\infty}(TM)$ is the vector-field expression of the control force, with $\tau^{a}_{M} \in\Gamma^{\infty}(T^{*}M)$ the base directions and $\tau_{G}\in\mathfrak{g}^{*}$ the fiber directions, and $h(\cdot,\cdot)$ calculated according to Appendix \ref{App:LCterms}. %
Note that \eqref{eq:ShapeDyn} and \eqref{eq:pp} describe the dynamics of the shape and the fiber coordinates, respectively, which are decoupled but related by \eqref{eq:xi}. 

Likewise, 
the reduced motion equations for an invariant constrained system $\Sigma_{\mathcal{D}}=\{Q,\mathbb{G},V,\mathcal{F},\mathcal{D}\}$ can be expressed as
\begin{align}
    \overset{\overline{\mathbb{M}}}{\nabla}_{\dot{r}(t)}\dot{r}(t) &= \frac{1}{2}\overline{\mathbb{M}}^{\sharp}(\bar{h}(\dot{\gamma}(t),\dot{\gamma}(t))) -\mathcal{P}_{M}\left(\mathrm{grad}V(r)\right) + \bar{f}_{u} , \label{eq:CnShapeDyn} \\ 
    \dot{\bar{p}}_{i}(t) &= \left\langle \frac{\partial l}{\partial \xi} ; [\xi , \underline{e}_{i} ] + \frac{\partial \underline{e}_{i}(r)}{\partial r^{i}} \dot{r}^{i} \right\rangle + \overline{\tau}_{G}, \label{eq:Cpp} \\
  \overline{\mathfrak{g}}\ni  \xi (t) &\triangleq g^{-1}\dot{g}(t) =  - \mathbb{A}(r)\dot{r}(t) + \bar{I}^{-1}(r) \bar{p}(t), \label{eq:Cxi}  
\end{align}
where $\bar{f}_{u} = \sum^{m}_{a=1} u^{a}(t) \overline{\mathbb{M}}^{\sharp}\left( \tau^{a}_{M} \right) \in \Gamma^{\infty}(TM)$ is the control force on the shape coordinates, $\mathcal{P}_{M}\vcentcolon TM \to \overline{\mathcal{H}}$ is the projection map, $\overline{\tau}_{G}\in\overline{\mathfrak{g}}^{*}$ denotes the external force in the restricted fiber directions, and $\bar{h}(\cdot,\cdot)$ is calculated as in Appendix \ref{App:LCterms}.  
Note that \eqref{eq:Cpp} is the time evolution of the generalized momentum \eqref{eq:pbdef} in its coordinate formulation \citep{bloch1996nonholonomic}. 

 \section{Controller Design}\label{sec:UnCtrl}
 \subsection{Problem formulation}
In this paper, we address the following two problems:

\noindent
{\bf Problem P1: Control of unconstrained systems}: For an invariant mechanical system $\Sigma=(Q,\mathbb{G},V,\mathcal{F})$ on a trivial principal fiber bundle $\Pi=(Q,G,\Phi,M)$ described by \eqref{eq:ShapeDyn}-\eqref{eq:xi} and a desired pose trajectory $g_{d}\vcentcolon \mathbf{I}\to G$ on the structure group $G$, which is assumed to be twice differentiable, design a control law for $f_u$ such that $g(t)\to g_d(t)$ asymptotically.

{\bf Problem P2: Control of constrained systems}: For a constrained invariant mechanical system $\Sigma=(Q,\mathbb{G},V,\mathcal{F},D)$ on a trivial principal fiber bundle $\Pi=(Q,G,\Phi,M)$ described by \eqref{eq:CnShapeDyn}-\eqref{eq:Cxi} and a desired pose trajectory $g_{d}\vcentcolon \mathbf{I}\to G$ on the structure group $G$, which is assumed to be twice differentiable, design a control law for $\bar{f_u}$ such that $g(t)\to g_d(t)$ asymptotically.

In the development of control laws, we make the following assumptions:
 \begin{assumption}[Mechanical connection]\label{Asm:MechConn}
    The local form of the mechanical connection $A(r)\vcentcolon T_{r}M \to \mathfrak{g}$ in \eqref{eq:p} for the unconstrained system, and $\mathbb A (r)\vcentcolon T_{r}M \to \bar{\mathfrak{g}}$ in \eqref{eq:pb} for the constrained system, has full rank, and its generalized inverse is denoted $(\cdot)^{\dagger}$.  
 \end{assumption}
  \begin{assumption}[Full actuation]\label{Asm:FullAct}
 $\mathrm{dim}(M) = m$, i.e., the dynamics of the shape coordinates is fully actuated. 
 \end{assumption}

In this section, we start with the problem P1. We will proceed with the design for the regulation problem based on the sliding mode strategy. Leveraging the Lie group structure, the tracking problem can be solved by properly defining a tracking error on the tangent bundle. Recall from Section \ref{sec:Model} that the motion equations of a mechanical system with symmetries on a trivial principal fiber bundle can be separated into two decoupled systems, one for the shape coordinates and the other for the pose coordinates. By exploiting this structural property, the reaching law will be designed on the shape space, which drives the pose trajectories to the sliding subgroup from an initial condition and then slides to the identity of the state manifold upon reaching the sliding subgroup to achieve the control objective.  

Toward this end, we use the class of kinematic controllers on the isomorphism $\mathcal{T}G\triangleq G\times \mathfrak{g}$ of the tangent bundle $TG=G\times T_{g}G$ introduced in \cite{espindola2025geometric} to map a group element $g$ to a vector in the Lie algebra, which will allow us to properly define a sliding subgroup immersed in the Lie group. 

\subsection{Kinematic control on a Lie group}\label{sec:KinCtrl}

Consider the kinematic system $\dot{g}(t)=g(t)\xi(t)$.  Taking the vector $\xi\in\mathfrak{g}$ as the control input, the kinematic control problem is to design a kinematic controller $\xi_u: G\to \mathfrak{g}$ to drive $g(t)$ to the group identity  $e$ asymptotically. 
We consider here the following class of kinematic controllers \citep{espindola2025geometric}. 
\begin{definition}[Kinematic controller]\label{def:KinCtrl}
A kinematic controller is 
a tuple $(\xi_{u}, V_{G})$,  where $V_{G}\vcentcolon G\to \mathbb{R}_{\geq 0}$ is a proper Morse function 
and the map $\xi_{u}\vcentcolon G\to\mathfrak{g}$ is such that
\begin{enumerate}\renewcommand{\theenumi}{\roman{enumi}}
    \item \label{def:kinCtrl1}$\xi_{u}(e) = 0$,
    \item \label{def:kinCtrl2} $\xi_{u}\left(g^{-1}\right) = -\xi_{u}(g)$, 
    \item \label{def:kinCtrl3} $\left\langle \mathrm{d}V_{G}(g); -g(t) \xi_{u}(g) \right\rangle <0$ for all $g(0)\in G\backslash\mathcal{O}$ and  $t\in\mathbf{I}$, where $\mathcal{O}\triangleq \left\{ g\in G\backslash \{ e\} \; |\; \xi_{u}(g)=0 \right\}$. 
    \item \label{def:kinCtrl4} $\left\langle \mathrm{d}V_{G}(g); -g(t)\xi_{u}(g) \right\rangle < -y(g)V_{G}(g)$, for all $g(0)\in\mathcal{U}$, $t\in\mathbf{I}$, where $y\vcentcolon\mathcal{U}\to \mathbb{R}_{> 0}$, and $\mathcal{U}\subset G\backslash \mathcal{O}$ is a neighborhood of $e$. 
\end{enumerate}
\end{definition}
A proper Morse function $V_{G}(g)$ is infinitely differentiable, $V_{G}(g)>0$ $\forall g\in G\backslash \{ e\}$, $\mathrm{d}V_{G}(g) = 0$ and $V_{G}(g)=0$  iff $g=e$. It measures the distance between an element $g$ and the identity $e$.  In addition, given a Riemannian metric $\mathbb{G}(g)$ on $G$, it satisfies $0< b_{1}\| \mathrm{d}V_{G} \|^{2}_{\mathbb{G}} (g)  \leq V_{G}(g) \leq b_{2}\|\mathrm{d}V_{G}\|^{2}_{\mathbb{G}}(g)$, for some $0<b_{1}\leq b_{2}$, for all $g\in \mathcal{U}\subset G$, $\mathcal{U}$ being a neighborhood of the identity $e$, which contains $e$ as the unique critical point \citep{maithripala2006almost}.

The properties (\ref{def:kinCtrl1}) and (\ref{def:kinCtrl2}) in Definition \ref{def:KinCtrl} allow for a proper definition of the sliding subgroup. The properties (\ref{def:kinCtrl3}) and (\ref{def:kinCtrl4}) ensure almost global asymptotic stability and local exponential stability of the equilibrium point $g=e$, respectively. Note that $\mathcal{O}$ represents the set of finite critical points for $V_{G}$ different from $e$. 
Moreover, it is nowhere dense, making $G\backslash \mathcal{O}$ an open and dense submanifold of $G$.

The following results from \cite{espindola2025geometric} establish that the state manifold $\mathcal{T}G=G\times \mathfrak{g} \simeq TG$ is a Lie group under a properly defined group operation, and there exists a sliding Lie subgroup immersed in the Lie group $\mathcal{T}G$ with the desired sliding property.

\begin{lem}[Lie group structure] 
Let $G$ be a Lie group with Lie algebra $\mathfrak{g}$. Then the following hold:

\noindent {\it (i)} $\mathcal{T}G=G\times \mathfrak{g}$ is a Lie group under the group operation 
$ \sigma_1*\sigma_2=(g_1g_2,\ \xi_1+\xi_2+\lambda\xi_u(g_1)+\lambda\xi_u(g_2)-\lambda \xi_u(g_1g_2)$, 
for $\sigma_1=(g_1,\xi_1)$, $\sigma_2=(g_2,\xi_2)\in \mathcal{T}G$, where $\lambda>0$ is a constant scalar;

\noindent {\it (ii)} Let  $G_{SL}\triangleq \big\{ \sigma = ( g , \xi ) \in \mathcal{T}G \; |\; \xi = -\lambda\xi_{u}(g)\big\} \subset \mathcal{T}G$, where $(\xi_{u},V_{G})$ is a kinematic controller. Then, $G_{SL}$ is a Lie subgroup; 

\noindent {\it (iii)} $G_{SL}$ is forward invariant, i.e., if $\sigma (t_{r})\in G_{SL}$ for some 
$t_r\geq 0$, then $\sigma (t)\in G_{SL}$, $\forall t\geq t_r$;

\noindent {\it (iv)} 
The equilibrium  $g=e$ 
of $\dot g= g \big(-\lambda \xi_u (g)\big)$ is almost globally asymptotically stable and locally exponentially stable.
\end{lem}

\subsection{Sliding modes control}\label{sec:SldCtrl}
We now develop a control law for the control force $f_{u}$ in the shape dynamics \eqref{eq:ShapeDyn} of the unconstrained system to drive the trajectory $(g(t), \xi (t))$ to the sliding subgroup $G_{SL}$ from almost any initial conditions. 

For this purpose, consider a curve in the shape space $r\vcentcolon \mathbf{I}\to M$, with $\dot{r}(t)\in T_{r(t)}M$, for all $t\in\mathbf{I}$, and define a {\it sliding} vector field on the curve $r(t)$ as
\begin{equation}
    s(t) =\dot{r}(t) + v(t) \in T_{r(t)}M, \label{eq:sTr} 
\end{equation}
where the vector field $v(t)\in \Gamma^{\infty}(TM)$ is defined as
\begin{equation}\label{eq:v}
    v (t) \triangleq -A^{\dagger}(r)\left( \lambda\xi_{u} (g) + I^{-1}(r)p(t) \right).
\end{equation}
Note that by \eqref{eq:xi} and \eqref{eq:sTr}, $s(t) = -A^{\dagger}(r)\left( \xi(t) +\lambda\xi_{u} (g) \right)$ and the  convergence of $s(t)\to 0$ imply $\xi(t) \to -\lambda \xi_{u}(g)$ under Assumption \ref{Asm:MechConn}. 

To make the sliding subgroup attractive and ensure the convergence of $\xi(t)\to -\lambda\xi_{u}(g)$, we propose the following controller
\begin{align}
    f_{u} &= -\overset{\mathbb{M}}{\nabla}_{\dot{r}(t)}v(t) - \frac{1}{2}\mathbb{M}^{\sharp}(h(\dot{\gamma}(t),\dot{\gamma}(t))) +\mathrm{grad}V(r) \notag\\
    &\quad - k_{s}s(t) + \mathbb{M}^{\sharp}\left(A^{T}(r)g^{-1}\mathrm{grad}V_{G}(g)\right). \label{eq:CtrlSld2}
\end{align}

\begin{lem}[Sliding modes control]\label{lem:SldBasedCtrl}
Let $\gamma (t) =  (g(t) , r(t)) \in Q$ with $\dot{\gamma}(t) = \left( g\xi (t) , \dot{r}(t)\right)\in T_{\gamma (t)}Q$ be a differentiable curve of an invariant mechanical system $\Sigma$ on a principal fiber bundle $\Pi$, whose reduced dynamics is described by \eqref{eq:ShapeDyn}-\eqref{eq:xi}. Then, under Assumptions \ref{Asm:MechConn}-\ref{Asm:FullAct} and given a kinematic controller $(\xi_{u}, V_{G})$, the control force \eqref{eq:CtrlSld2} ensures the convergence $(g(t),\xi(t)) \to (e,0)$ asymptotically for all $(g(0),\xi (0))\in G\backslash\mathcal{O}\times \mathfrak{g}$, and exponentially for all $(g(0),\xi (0))\in \mathcal{U}\times \mathfrak{g}$.
\end{lem}
\begin{pf} 
The time evolution of $s(t)$ along the curve $r(t)$ is calculated by
\begin{equation*}
    \overset{\mathbb{M}}{\nabla}_{\dot{r}(t)}s(t) = \overset{\mathbb{M}}{\nabla}_{\dot{r}(t)}\dot{r}(t) + \overset{\mathbb{M}}{\nabla}_{\dot{r}(t)}v(t).
\end{equation*}
By \eqref{eq:ShapeDyn} and \eqref{eq:cConn} it yields 
\begin{align}\label{eq:sDyn}
    \overset{\mathbb{M}}{\nabla}_{\dot{r}(t)}s(t) &=  \overset{\mathbb{M}}{\nabla}_{\dot{r}(t)}v(t) + \frac{1}{2}\mathbb{M}^{\sharp}(h(\dot{\gamma}(t),\dot{\gamma}(t))) -\mathrm{grad}V(r) \notag\\
    &\quad  + f_{u},
\end{align}
which describes the dynamics of the vector $\xi(t) + \lambda \xi_{u}(g) \in \mathfrak{g}$, expressed in the shape space $M$ through $s(t)$. 

Applying control force \eqref{eq:CtrlSld2} to the dynamical system \eqref{eq:sDyn} leads to
\begin{equation*}
    \overset{\mathbb{M}}{\nabla}_{\dot{r}(t)}s(t) = - k_{s}s(t)  + \mathbb{M}^{\sharp}\left(A^{T}(r)g^{-1}\mathrm{grad}V_{G}(g)\right) .
\end{equation*}

Consider the positive definite function  $W_{G}\vcentcolon G\times TM \to \mathbb{R}_{\geq 0}$ defined as $W_{G}(g,s) \triangleq V_{G}(g) + \frac{1}{2}\mathbb{M}(s,s)$, which satisfies $W_{G}(g,s) = 0$ iff $(g,s) = (e,0)$  $\forall (g,s\in G\backslash\mathcal{O}\times TM$. The time evolution of $W_{G}$ is given by
\begin{align*}
    \frac{\mathrm{d}}{\mathrm{d}t}&W_{G}(g,s) = \left\langle \mathrm{d}V_{G}(g); g \xi (t) \right\rangle + \mathbb{M}\left( s(t) , \overset{\mathbb{M}}{\nabla}_{\dot{r}(t)}s(t) \right) \\
    = &\left\langle \mathrm{d}V_{G}(g); g \xi (t) \right\rangle \\
    &+ \mathbb{M} \left( s(t) , - k_{s}s(t) + \mathbb{M}^{\sharp}\left(A^{T}(r)g^{-1}\mathrm{grad}V_{G}(g) \right) \right) ,
    \end{align*}
adding and subtracting $\lambda\xi_{u}(g)$ to $\xi(t)$, and using \eqref{eq:sTr} gives 
    \begin{align*}
    &\frac{\mathrm{d}}{\mathrm{d}t}W_{G}(g,s) =   \\
    &\quad \left\langle \mathrm{d}V_{G}(g); g \left(\xi (t) + \lambda\xi_{u}(g)\right) \right\rangle + \left\langle \mathrm{d}V_{G}(g); -\lambda g\xi_{u}(g) \right\rangle \\
    &\quad + \mathbb{M} \left( s(t) , - k_{s}s(t) + \mathbb{M}^{\sharp}\left(A^{T}(r)g^{-1}\mathrm{grad}V_{G}(g) \right) \right)\\
    &= \left\langle \mathrm{d}V_{G}(g); g\left( - A(r)s(t)\right) \right\rangle + \left\langle \mathrm{d}V_{G}(g); -\lambda g\xi_{u}(g) \right\rangle \\
    &\quad + \mathbb{M} \left( s(t) , - k_{s}s(t) \right) + \left\langle A^{T}(r)g^{-1}\mathrm{grad}V_{G}(g) ; s(t)  \right\rangle \\    
    &= \left\langle \mathrm{d}V_{G}(g); -\lambda g\xi_{u}(g) \right\rangle -k_{s}\mathbb{M} \left( s(t) , s(t) \right).
\end{align*}
Therefore, according to Definition \ref{def:KinCtrl}-$(iii)$, 
it yields $\frac{\mathrm{d}}{\mathrm{d}t}W_{G}(g,s) < 0$, for all $(g(0),s(0))\in G\backslash\mathcal{O}\times TM$. This ensures asymptotically the convergence $(g(t),\xi(t)) \to (e,0)$ for all $(g(0),\xi (0))\in G\backslash\mathcal{O}\times \mathfrak{g}$. 

Now, for $(g(0),s(0))\in \mathcal{U}\times TM$, by Definition \ref{def:KinCtrl}-$(iv)$ it holds $\frac{\mathrm{d}}{\mathrm{d}t}W_{G}(g,s) = -y(g)V_{G}(g)-k_{s}\mathbb{M} \left( s(t) , s(t) \right) \leq -\mathrm{min}\left\{ \lambda \underline{y},k_{s}\right\} W_{G}(g,s)$, with $\underline{y} = \mathrm{min}\{ y(g)\},\; \forall g\in\mathcal{U}$. This shows the exponential convergence of $(g(t),\xi(t)) \to (e,0)$ for $(g(0),s(0))\in \mathcal{U}\times TM$.
 \hfill$\blacksquare$
\end{pf}

\subsection{Tracking Control}\label{sec:TckCtrl}
We consider in this subsection the problem of tracking a desired pose trajectory $g_{d}\vcentcolon \mathbf{I}\to G$, which is assumed to be twice differentiable, on the structure group $G$ of a trivial principal fiber bundle. The desired velocity satisfies the kinematic equation $\dot{g}_{d}(t)=g_{d}(t)\xi_{d}(t)$ for a desired body velocity  $\xi_{d}\vcentcolon\mathbf{I}\to\mathfrak{g}$. 
Then, two different errors can be defined as elements of the group, either by a left-translation $g^{L}_{e} = g^{-1}_{d}(t)g(t)$ (left error) or by a right-translation $g^{R}_{e} = g(t)g^{-1}_{d}(t)$ (right errors). Either $g^L_{e}\to e$ or $g^R_{e}\to e$ represents the control objective of the tracking problem. We consider here the left error $g^L_{e}$ and denote it as $g_{e}$. Its time evolution is calculated as
\begin{align}
    \dot{g}_{e}(t) &= g_{e}(t)\xi_{e}(t), \label{eq:gep}\\
    \xi_{e}(t) & = \xi(t)-\mathrm{Ad}_{g^{-1}_{e}}\xi_{d}(t) .\label{eq:xie}
\end{align}
Note that kinematics \eqref{eq:gep} has the same structure as that analyzed in Section \ref{sec:KinCtrl}, with $\xi_{e}(t) \in\mathfrak{g}$ as the control input. Therefore, as discussed in the last subsection, for each curve $r\vcentcolon \mathbf{I}\to M$ in the shape space, define the following sliding vector field on the curve $r(t)$
\begin{equation}
    \varsigma(t) = \dot{r}(t) + w(t) \in T_{r(t)}M, \label{eq:s2}  
\end{equation}
where the vector field $w(t)\in\Gamma^{\infty}(TM)$ is defined as
\begin{equation}\label{eq:w}
    w(t) \triangleq -A^{\dagger}(r)\left( \lambda\xi_{u}(g_{e}) -\mathrm{Ad}_{g^{-1}_{e}}\xi_{d}(t) + I^{-1}(r)p(t) \right).
\end{equation}
In view of \eqref{eq:xi} and \eqref{eq:xie}, 
$ \varsigma(t)= -A^{\dagger}(r)\left( \xi_{e}(t) + \lambda\xi_{u}(g_{e})\right)$.
    Then according to \eqref{eq:CtrlSld2}, the following tracking controller for the system \eqref{eq:ShapeDyn}-\eqref{eq:xi} is proposed:
\begin{align}
   f_{u} &= -\overset{\mathbb{M}}{\nabla}_{\dot{r}(t)}w(t) - \frac{1}{2}\mathbb{M}^{\sharp}(h(\dot{\gamma}(t),\dot{\gamma}(t))) +\mathrm{grad}V(r) \notag\\
    &\quad - k_{s}\varsigma(t) + \mathbb{M}^{\sharp}\left(A^{T}(r)g^{-1}_{e}\mathrm{grad}V_{G}(g_{e})\right). \label{eq:TckCtrl}
\end{align}

\begin{thm}[Sliding based tracking control]\label{thm:TckCtrl}
Let $\Sigma$ be an invariant mechanical system described by \eqref{eq:ShapeDyn}-\eqref{eq:xi}, evolving on a trivial principal fiber bundle $\Pi$, and $(\xi_{u}, V_{G})$ be a kinematic controller on the structure group $G$. Then, under Assumptions \ref{Asm:MechConn}-\ref{Asm:FullAct} the control force \eqref{eq:TckCtrl} achieves the convergence $(g(t),\xi (t)) \to (g_{d}(t),\xi_{d} (t))$ asymptotically $\forall \; (g(0),\xi (0))\in G\backslash \mathcal{O}\times\mathfrak{g}$, and exponentially for all $(g(0),\xi (0))\in \mathcal{U}\times \mathfrak{g}$. 
\end{thm}
\begin{pf}
   By the system dynamics \eqref{eq:ShapeDyn}-\eqref{eq:xi} and the control law \eqref{eq:TckCtrl} the dynamics of the closed loop is 
    \begin{equation*}
        \overset{\mathbb{M}}{\nabla}_{\dot{r}(t)}\varsigma (t) = -k_{s}\varsigma(t) + \mathbb{M}^{\sharp}\left( A^{T}(r)g^{-1}_{e} \mathrm{grad}V_{G}(g_{e}) \right).  
    \end{equation*}
     The rest of the proof follows from the same lines as in the proof of Lemma \ref{lem:SldBasedCtrl} using the positive definite function $W_{G}(g_{e} (t),\varsigma (t)) = V_{G}(g_{e}) + \frac{1}{2}\mathbb{M}(\varsigma (t),\varsigma (t))$. 
    \hfill$\blacksquare$
\end{pf}

\section{Control of Constrained Systems}\label{sec:CnCtrl}
We now proceed to design a sliding mode controller for an invariant constrained system $\Sigma_{\mathcal{D}}=\{Q,\mathbb{G},V,\mathcal{F},\mathcal{D}\}$ on a trivial principal fiber bundle \eqref{eq:CnShapeDyn}-\eqref{eq:Cxi}. Similarly to the unconstrained case, we first address stabilizing $g(t)$ to the identity of the Lie group $G$, assuming the existence of a kinematic controller $\xi_u:G\to \bar{\mathfrak{g}}$ with the properties $(i)-(iv)$ in Definition \ref{def:KinCtrl}, we solve the tracking to a desired pose trajectory on $\overline{\mathcal{T}G}\triangleq G\times \bar{\mathfrak{g}}$ by leveraging the Lie group structure while ensuring the nonholonomic constraints. 

\subsection{Regulation of constrained systems}
The kinematic controller $(\xi_{u} , V_{G})$ is constrained on the distribution $D$, with 
$\xi_{u}\vcentcolon G \to \bar{\mathfrak{g}}$. Assume that this kinematic controller fulfills the properties $(i)-(iv)$ in Definition \ref{def:KinCtrl},  which concerns a controllability condition \citep{bullo2002controllability,cortes2002simple}, then the same kinematic controller can achieve stabilizing $g$ to the identity in the kinematics $\dot g=g \xi$ under the kinematic control. 
For the reaching stage, since the velocity error $\xi_e=\xi\in \bar{\mathfrak{g}}$ for the regulation problem, the reaching controller \eqref{eq:TckCtrl} constrained on $D$ achieves the goal.

\begin{lem}[Constrained sliding based controller]\label{lem:nhSldBsCtrl}
Let $(\xi_{u},V_{G})$ be a constrained 
kinematic controller, i.e., $\xi_{u}: G\to \bar{\mathfrak{g}}$. Then under Assumptions \ref{Asm:MechConn}-\ref{Asm:FullAct}, the nonholonomic system $\Sigma_{\mathcal{D}}$ with reduced dynamics \eqref{eq:CnShapeDyn} in closed loop with controller
    \begin{align}
    \bar{f}_{u} &= -\overset{\overline{\mathbb{M}}}{\nabla}_{\dot{r}(t)}\bar{v}(t) - \frac{1}{2}\overline{\mathbb{M}}^{\sharp}(\bar{h}(\dot{\gamma}(t),\dot{\gamma}(t))) +\mathcal{P}\left(\mathrm{grad}V(r)\right) \notag\\
    &\quad - k_{s}\bar{s}(t) + \overline{\mathbb{M}}^{\sharp}\left(\mathbb{A}^{T}(r)g^{-1}\mathrm{grad}V_{G}(g)\right) , \label{eq:nhCtrlSld2} \\
        \bar{v}(t) &= -\mathbb{A}^{\dagger}(r)\left( \lambda\xi_{u} (g) + \bar{I}^{-1}(r)\bar{p}(t) \right) ,\notag\\
        \bar{s}(t) &= \dot{r}(t) + \bar{v}(t), \notag
    \end{align}
where $k_{s}>0$ is a constant scalor control gain,  
ensures the convergence $(g(t),\xi (t))\to (e,0)$ asymptotically for all $(g(0),\xi(0))\in G\backslash\mathcal{O}\times \overline{\mathfrak{g}}$, and exponentially for all $(g(0),\xi(0))\in \mathcal{U}\times \overline{\mathfrak{g}}$. 
\end{lem}
\begin{pf}
By the reduced dynamics \eqref{eq:CnShapeDyn} and the controller \eqref{eq:nhCtrlSld2} the closed-loop system is 
\begin{equation*}
     \overset{\overline{\mathbb{M}}}{\nabla}_{\dot{r}(t)}\bar{s}(t) = -k_{s}\bar{s}(t) + \overline{\mathbb{M}}^{\sharp}\left( \mathbb{A}^{T}(r)g^{-1} \mathrm{grad}V_{G}(g)\right),
\end{equation*}
The rest of the proof follows the same lines as in the proof of Lemma \ref{lem:SldBasedCtrl} with the positive definite function $\overline{W}_{G}(g,\bar{s}) = V_{G}(g) + \frac{1}{2}\overline{\mathbb{M}}(\bar{s},\bar{s})$ instead. 
    \hfill$\blacksquare$
\end{pf}

\subsection{Tracking control for constrained systems}
The major difficulty encountered in developing tracking controls for constrained systems is that for $\xi(t),\xi_{d}(t)\in\overline{\mathfrak{g}}$, the tracking error $\xi_{e}(t) = \xi(t)-\mathrm{Ad}_{g^{-1}_{e}}\xi_{d}(t)$ in kinematics \eqref{eq:gep} generally does not hold the nonholonomic constraints due to adjoint operation. 
To address this issue, we make the following additional assumption.

\begin{assumption}[Morse function]\label{Asm:nhKinCtrl}
For the Morse function $V_G$ of a nonholonomic kinematic controller $(\xi_{u}, V_{G})$, there exists a constant $\gamma_{G}>0$ such that 
  \begin{equation}\label{eq:nhCondition}
    V_{G}(g)\geq\gamma_{G} \left\|\mathrm{Ad}_{e}-\mathrm{Ad}_{g^{-1}}\right\|^{2}_{\bar{I}},  \ \forall g\in\mathcal{U}_{e}, 
  \end{equation}
where $\mathcal{U}_{e}\subset G$ is a neighborhood of identity $e$. 
 \end{assumption}
 
Define a nonholonomic sliding vector field on the curve $r(t)$ as
\begin{equation}\label{eq:nhs2}
    \bar{\varsigma} (t)  \triangleq \dot{r}(t) + \overline{w}(t), 
\end{equation}
where the vector field $\overline{w}\in\Gamma^{\infty}(TM)$ on the shape space is given by 
\begin{equation}\label{eq:nhw}
    \overline{w}(t) \triangleq -\mathbb{A}^{\dagger}(r)\left( \lambda\xi_{u}(g_{e}) -\xi_{d}(t) + \bar{I}^{-1}(r)\bar{p}(t)  \right). 
\end{equation}
By \eqref{eq:nhs2} and \eqref{eq:nhw} it yields $\bar{\varsigma} = -\mathbb{A}^{\dagger}(r)\left( \xi(t)+ \lambda\xi_{u}(g_{e}) \right)$. Therefore,  in view of \eqref{eq:nhCtrlSld2} a tracking control law for an invariant nonholonomic mechanical system $\Sigma_{\mathcal{D}}$ is proposed as
 \begin{align}
        \bar{f}_{u} &= -\overset{\overline{\mathbb{M}}}{\nabla}_{\dot{r}(t)}\overline{w}(t) - \frac{1}{2}\overline{\mathbb{M}}^{\sharp}(\bar{h}(\dot{\gamma}(t),\dot{\gamma}(t))) +\mathcal{P}\left(\mathrm{grad}V(r)\right) \notag\\
    &\quad - k_{s}\bar{\varsigma}(t) + \overline{\mathbb{M}}^{\sharp}\left(\mathbb{A}^{T}(r)g^{-1}_{e}\mathrm{grad}V_{G}(g_{e})\right) . \label{eq:nhTckCtrl}
\end{align}

\begin{thm}[Nonholonomic tracking control]\label{thm:nhTckCtrl}
Consider the nonholonomic mechanical system $\Sigma_{\mathcal{D}}$ with symmetries on a principal fiber bundle $\Pi$, described by \eqref{eq:CnShapeDyn}-\eqref{eq:Cxi}, and a nonholonomic kinematic controller $(\xi_{u}, V_{G})$ on the structure group $G$. Under Assumptions \ref{Asm:MechConn}-\ref{Asm:FullAct},  and \ref{Asm:nhKinCtrl}, the nonholonomic control force \eqref{eq:nhTckCtrl} achieves exponentially the convergence of $(g(t),\xi (t)) \to (g_{d}(t),\xi_{d} (t))$ for all $(g(0),\xi(0))\in \mathcal{U}\times \bar{\mathfrak{g}}$. 
\end{thm}
\begin{pf}
   By \eqref{eq:CnShapeDyn} and \eqref{eq:nhTckCtrl}, the covariant derivative of the vector field \eqref{eq:nhs2} along the curve $r(t)\in M$ is
    \begin{equation*}
        \overset{\overline{\mathbb{M}}}{\nabla}_{\dot{r}(t)}\bar{\varsigma} (t) = -k_{s}\bar{\varsigma}(t) + \overline{\mathbb{M}}^{\sharp}\left( \mathbb{A}^{T}(r)g^{-1}_{e} \mathrm{grad}V_{G}(g_{e}) \right) ,
    \end{equation*}
which has an equilibrium point $(g_{e},\bar{\varsigma}) = (e,0)$. To prove its stability, consider the positive definite function $\overline{W}_{G}(g_{e} (t),\bar{\varsigma} (t)) = V_{G}(g_{e}) + \frac{1}{2}\overline{\mathbb{M}}(\bar{\varsigma} (t),\bar{\varsigma} (t))$. 
Its time evolution is given by
    \begin{align*}
    &\frac{\mathrm{d}}{\mathrm{d}t}\overline{W}_{G}(g_{e},\bar{\varsigma}) =  
    \left\langle \mathrm{d}V_{G}(g_{e}); g_{e}\xi_{e} (t) \right\rangle + \overline{\mathbb{M}}\big( \bar{\varsigma}(t) , \overset{\overline{\mathbb{M}}}{\nabla}_{\dot{r}(t)}\bar{\varsigma}(t) \big) \\
     =& \left\langle \mathrm{d}V_{G}(g_{e}); g_{e} \left( \xi (t) - \mathrm{Ad}_{g^{-1}_{e}}\xi_{d}(t)\right) \right\rangle \\
     &+ \overline{\mathbb{M}} \left( \bar{\varsigma}(t) , - k_{s}\bar{\varsigma}(t) + \overline{\mathbb{M}}^{\sharp}\left(\mathbb{A}^{T}(r)g^{-1}_{e}\mathrm{grad}V_{G}(g_{e}) \right) \right) ,
    \end{align*}
    where \eqref{eq:xie} was used. Adding and subtracting $\lambda\xi_{u}(g_{e})-\xi_{d}(t)$ to $\xi(t)$, and using \eqref{eq:nhs2} yields
    \begin{align*}
    &\frac{\mathrm{d}}{\mathrm{d}t}\overline{W}_{G}(g_{e},\bar{\varsigma}) =  
    \left\langle \mathrm{d}V_{G}(g_{e}); g_{e} \left( \xi (t) -\xi_{d}(t) + \lambda\xi_{u}(g_{e})  \right) \right\rangle \notag\\
    &\quad + \left\langle \mathrm{d}V_{G}(g_{e}); g_{e}\left( - \lambda\xi_{u}(g_{e}) + \xi_{d}(t) - \mathrm{Ad}_{g^{-1}_{e}}\xi_{d}(t) \right)\right\rangle \\
    &\quad + \overline{\mathbb{M}} \left( \bar{\varsigma}(t) , - k_{s}\bar{\varsigma}(t) + \overline{\mathbb{M}}^{\sharp}\left( \mathbb{A}^{T}(r)g^{-1}_{e}\mathrm{grad}V_{G}(g_{e}) \right) \right) \\
    &= \left\langle \mathrm{d}V_{G}(g_{e}); g_{e}\left( - \mathbb{A}(r)\bar{\varsigma}(t)\right) \right\rangle + \left\langle \mathrm{d}V_{G}(g_{e}); -\lambda  g_{e} \xi_{u}(g_{e})\right\rangle  \\
    &\quad  + \left\langle \mathrm{d}V_{G}(g_{e}); g_{e} \left( \xi_{d}(t) - \mathrm{Ad}_{g^{-1}_{e}}\xi_{d}(t) \right)\right\rangle \\
    &\quad + \overline{\mathbb{M}} \left( \bar{\varsigma}(t) , - k_{s}\bar{\varsigma}(t) \right) + \left\langle \mathbb{A}^{T}(r)g^{-1}_{e}\mathrm{grad}V_{G}(g_{e}) ; \bar{\varsigma}(t)  \right\rangle  \\    
    &= -\lambda\left\langle \mathrm{d}V_{G}(g_{e});  g_{e} \xi_{u}(g_{e})\right\rangle -k_{s}\overline{\mathbb{M}} \left( \bar{\varsigma}(t) , \bar{\varsigma}(t) \right)  \\
    &\quad + \left\langle \mathrm{d}V_{G}(g_{e}); g_{e} \left( \xi_{d}(t) - \mathrm{Ad}_{g^{-1}_{e}}\xi_{d}(t) \right)\right\rangle \notag\\
    &\leq -\lambda\left\langle \mathrm{d}V_{G}(g_{e});  g_{e} \xi_{u}(g_{e})\right\rangle -k_{s}\overline{\mathbb{M}} \left( \bar{\varsigma}(t) , \bar{\varsigma}(t) \right)  \notag\\
    &\quad +\|\mathrm{d}V_{G}\|_{\mathbb{I}}\left\|\mathrm{Ad}_{e}-\mathrm{Ad}_{g^{-1}_{e}}\right\|_{\bar{I}}\|\xi_{d}\|_{\bar{I}},
\end{align*}
where $\mathbb{I}(g)$ is the Riemannian metric on $G$ induced by the tensor $\bar{I}(r)$ on $\bar{\mathfrak{g}}$. Besides, without loss of generality, let $\mathcal{U}\subseteq \mathcal{U}_{e}$ in Assumption \ref{Asm:nhKinCtrl} with $\mathcal{U}$ given in Definition \ref{def:KinCtrl}. Then Definition \ref{def:KinCtrl}-$(iv)$ and Eq. \eqref{eq:nhCondition} yield
\begin{align*}
    &\frac{\mathrm{d}}{\mathrm{d}t}\overline{W}_{G}(g_{e},\bar{\varsigma}) \leq -\lambda y(g_{e})V_{G}(g_{e})  -k_{s}\overline{\mathbb{M}} \left( \bar{\varsigma}(t) , \bar{\varsigma}(t) \right)  \notag   \\
    &\vspace{10pt} +\frac{1}{\sqrt{b_{1}\gamma_{G}}}\|\xi_{d}\|_{\bar{I}}V_{G}(g_{e}) \\
    =& -\left( \lambda y(g_{e}) - \frac{1}{\sqrt{b_{1}\gamma_{G}}}\|\xi_{d}\|_{\bar{I}} \right) V_{G}(g_{e}) \notag\\
    &- k_{s}\overline{\mathbb{M}} \left( \bar{\varsigma}(t) , \bar{\varsigma}(t) \right). 
\end{align*}
Let  $\lambda \triangleq \lambda' \frac{\sup{ \left\{ \|\xi_{d}\|_{\bar{I}} \;|\; \xi_{d}\in \bar{\mathfrak{g}} \right\}  }}{\inf{\left\{ y(g_{e}) \;|\; g_{e}\in \mathcal{U} \right\} } \sqrt{b_{1}\gamma_{G}}}$, with $\lambda' > 1$, it has
\begin{align*}
    \frac{\mathrm{d}}{\mathrm{d}t}\overline{W}_{G}(g_{e},\bar{\varsigma}) \leq -\lambda_{\overline{W}_{G}} \overline{W}_{G}(g_{e},\bar{\varsigma}).
\end{align*}
where $\lambda_{\overline{W}_{G}}=\min{ \left\{ \frac{\sup{ \left\{ \|\xi_{d}\|_{\bar{I}} \;|\; \xi_{d}\in \bar{\mathfrak{g}} \right\}  }}{ \sqrt{b_{1}\gamma_{G}}} \left( \lambda' - 1 \right) ,\;  2k_{s} \right\} }>0$. Therefore, $(g_{e}(t),\bar{\varsigma}(t))$ converges to $(e,0)$ exponentially. 
    \hfill$\blacksquare$
\end{pf}

\begin{rem}[Tracking for constrained systems]\label{rem:KinTckCtrl}
To address the issue that the tracking error state $\sigma_e=(g_e,\xi_e)\in {\mathcal{T}G}$  generally does not belong to $\overline{\mathcal{T}G}$, 
the error velocity $\xi_e=\xi-\mathrm{Ad}_{g^{-1}_{e}}\xi_{d}\in \mathcal{T}G$ is projected to $\overline{\mathcal{T}G}$ by $\xi-\xi_d\in \overline{\mathcal{T}G}$, and 
an auxiliary error state $\bar{\sigma}_e=(g_e,\xi-\xi_d)\in \overline{\mathcal{T}G}$ is introduced. The distance between these two velocity errors is bounded by $\|\xi_e-(\xi-\xi_d)\|_{\overline{I}}=\|\xi_d-\mathrm{Ad}_{g^{-1}_{e}}\xi_{d}\|_{\overline{I}} \leq \|\mathrm{Ad}_{e}-\mathrm{Ad}_{g^{-1}_{e}}\|_{\overline{I}}\|\xi_d\|_{\overline{I}}$. Therefore, Assumption \ref{Asm:nhKinCtrl} is about finding a constant such that $\gamma_G \|\mathrm{Ad}_{e}-\mathrm{Ad}_{g^{-1}_{e}}\|_{\overline{I}}$ is dominated by the Morse function $V_G(g)$ for all $g\in G$, as illustrated in Fig. \ref{fig:nhTckCtrl} as well as in the unicycle mobile robot in Section \ref{sec:Examp} .

\begin{figure}
\begin{center}
\includegraphics[scale=0.25,trim = 0mm 0mm 0mm 0mm]{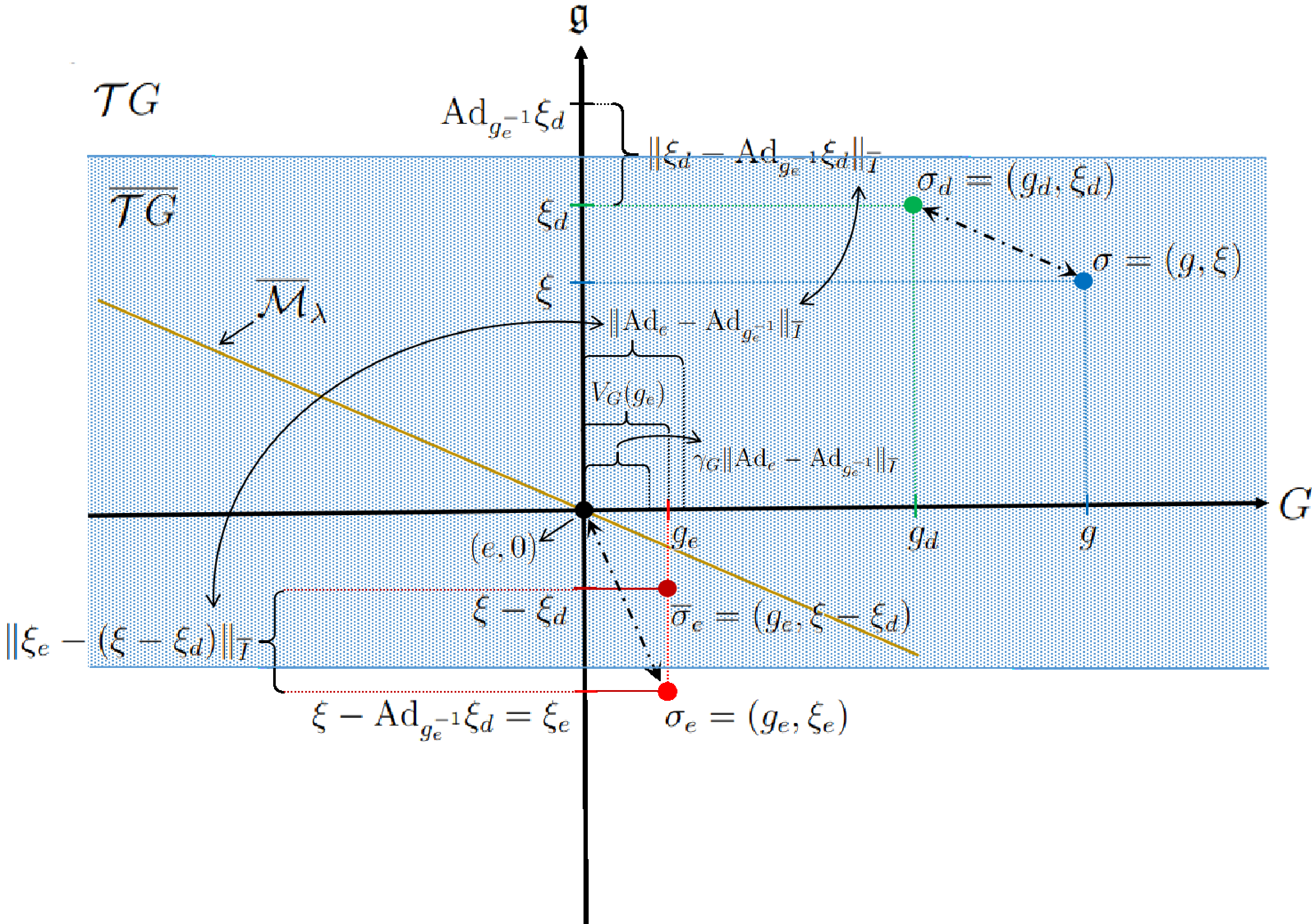}
\caption{Concept sketch of Assumption \ref{Asm:nhKinCtrl}. (1) The control objective is to make the state (blue dot) $\sigma=(g,\xi)\in \overline{\mathcal{T}G}$ (shaded area) to track a desired state (green dot) $\sigma_d=(g_d,\xi_d)\in \overline{\mathcal{T}G}$. 
(2) Leveraging the structure of the Lie group, a tracking error state $\sigma_e=(g_e,\xi_e)\in {\mathcal{T}G}$ is defined, which in general does not belong to $\overline{\mathcal{T}G}$.  
(3) An auxiliary error state (dark red) $\bar{\sigma}_e=(g_e,\xi-\xi_d)\in \overline{\mathcal{T}G}$ is introduced, by projecting the error velocity $\xi_e\in \mathcal{T}G$ to $\overline{\mathcal{T}G}$. 
(4) The distance between velocity errors is bounded by $\|\xi_e-(\xi-\xi_d)\|_{\overline{I}}=\|\xi_d-\mathrm{Ad}_{g^{-1}_{e}}\xi_{d}\|_{\overline{I}} \leq \|\mathrm{Ad}_{e}-\mathrm{Ad}_{g^{-1}_{e}}\|_{\overline{I}}\|\xi_d\|_{\overline{I}}$. 
(5) For a given kinematic controller $(\xi_u, V_G)$, 
Assumption \ref{Asm:nhKinCtrl} consists of finding a positive constant $\gamma_G$ such that $\gamma_G \|\mathrm{Ad}_{e}-\mathrm{Ad}_{g^{-1}_{e}}\|_{\overline{I}}$ is dominated by the Morse function $V_G(g)$ for all $g\in G$.
}
\label{fig:nhTckCtrl}                                 
\end{center}                                 
\end{figure}

\end{rem}

\section{Application Examples}\label{sec:Examp}
We apply the previous results to a spacecraft actuated by three reaction wheels with no constraints and 
a unicycle mobile robot actuated by two wheels under a nonholonomic constraint. The former is a generalization of attitude control in, for instance,   \cite{bullo1999tracking,espindola2025geometric} with actuator dynamics, and the latter was considered in \cite{tayefi2019logarithmic} using a backstepping design technique. 

\subsection{Attitude tracking of a spacecraft actuated by reaction wheels}
The attitude of a rigid spacecraft is fully represented by an element $R\in SO(3)$, with $SO(3)=\{ R\in\mathbb{R}^{3\times 3} \; | \; \mathrm{det}(R)=1, \; R^{T}R = RR^{T} = I_{3} \}$ the special orthogonal group of rotation matrices in a 3D space, where $I_{3}$ is the identity matrix $3\times 3$. That is, $R$ expresses the orientation of a body-fixed reference frame with respect to an inertial reference frame. In addition, $SO(3)$ forms a Lie group with the operation of the group the usual matrix product, the identity element $I_{3}$, and the inverse element the transposed matrix. Its Lie algebra is the set of skew-symmetric matrices given by $\mathfrak{so}(3)=\{ A\in\mathbb{R}^{3}\;| \; A=-A^{T} \}$, which is isomorphic to $\mathbb{R}^{3}$, $\mathfrak{so}(3)\simeq \mathbb{R}^{3}$. The adjoint operator is the cross product in $\mathbb{R}^{3}$, $\mathrm{ad}_{\xi ^{\wedge}}\zeta^{\wedge} = (\xi \times \zeta)^{\wedge}$, for all $\xi,\zeta\in\mathbb{R}^{3}$, where $(\cdot)^{\wedge}\vcentcolon \mathbb{R}^{3}\to \mathfrak{so}(3)$ is the {\it wedge} map with the inverse map $(\cdot)^{\vee}\vcentcolon \mathfrak{so}(3) \to \mathbb{R}^{3}$. Therefore, for a curve $R\vcentcolon\mathbf{I}\to SO(3)$, the body angular velocity $\Omega\vcentcolon \mathbf{I}\to \mathbb{R}^{3}$ is given by 
\begin{equation*}
    \Omega(t) ^{\wedge} \triangleq R^{T}(t)\dot{R}(t) = \left[ \begin{array}{ccc}
         0 & -\Omega_{3}(t) & \Omega_{2}(t)  \\
         \Omega_{3}(t) & 0 & -\Omega_{1}(t) \\
         -\Omega_(t) & \Omega_{1}(t) & 0
    \end{array} \right]. 
\end{equation*}
The spacecraft is actuated by three reaction wheels, each aligned to the principal axis of the body frame attached to the center of mass of the spacecraft. The rotation of the wheel with respect to the body frame is denoted by $\phi = (\phi_{1},\phi_{2},\phi_{3})\in \mathbb{S}^{1}\times\mathbb{S}^{1}\times \mathbb{S}^{1}$. Therefore, the configuration manifold is $Q=SO(3)\times \mathbb{T}^{3}$, with $\mathbb{T}^{3} = \mathbb{S}^{1}\times\mathbb{S}^{1}\times \mathbb{S}^{1}$. A group action $\Phi(R_{a},q)=(R_{a}R,\phi)\in Q$, $\forall q=(R,\phi)\in Q$ is defined with $R_{a}\in SO(3)$. 
Therefore, $(Q,SO(3),\Phi , \mathbb{T}^{3})$ forms a trivial principal fiber bundle.

A reduced metric can be directly proposed as
\begin{align*}
\mathcal{G}\big((\Omega , \dot{\phi}),(\Omega , \dot{\phi})\big) &= (J+J_{\phi})(\Omega , \Omega) + J_{\phi}(\Omega , \dot{\phi}) + J_{\phi}( \dot{\phi}, \Omega ) \notag\\
&\quad+ J_{\phi}(\dot{\phi} , \dot{\phi}) ,
\end{align*}
where $[J] = [J]^{T}\in\mathbb{R}^{3\times 3}$ is the moment of inertia of the spacecraft and $[J_{\phi}]=\mathrm{diag}\{J_{\phi,1},J_{\phi,2},J_{\phi,3}\} \in\mathbb{R}^{3\times 3}$ is the moment of inertia of the wheels, both are constant, symmetric and positive definite. Therefore, the reduced Lagrangian of the system is given by the kinetic energy associated to the metric $\mathcal{G}$ as $l(\phi,\Omega,\dot{\phi}) = \frac{1}{2}\mathcal{G}\big( (\Omega , \dot{\phi}),(\Omega, \dot{\phi}))\big)$, expressed in matrix form as
\begin{equation*}
    l\big(\phi,\Omega,\dot{\phi}\big) = \frac{1}{2}\left[ \Omega^{T}(t) \;\dot{\phi}^{T}(t) \right]\left[ \begin{array}{cc}
         J+J_{\phi} & J_{\phi}  \\
         J_{\phi}& J_{\phi}
    \end{array} \right] \left[ \begin{array}{c}
         \Omega(t)\\
         \dot{\phi} (t)
    \end{array} \right].
\end{equation*}
Thus it follows from \eqref{eq:rMetric} that $I=J+J_{\phi}$, $m = J_{\phi}$ and $IA=J_{\phi}$. Therefore, $A = (J+J_{\phi})^{-1}J_{\phi}$, which is invertible with $A^{-1} = J^{-1}_{\phi}(J+J_{\phi})$. Moreover, the block diagonalization \eqref{eq:rDMetric} of $\mathcal{G}$ results in $\mathbb{M}=J_{\phi} - J_{\phi}(J+J_{\phi})^{-1}J_{\phi}$. 
By \eqref{eq:ShapeDyn}-\eqref{eq:xi} the reduced equations of motion are 
\begin{align*} 
    \ddot{\phi}(t) &= \frac{1}{2}\mathbb{M}^{-1}\left(h\left(\Omega,\dot{\phi}\right)\right)+ f_{u}, \notag\\
    h\left( \Omega,\dot{\phi} \right) &= -2\left( (J+J_{\phi})^{-1}J_{\phi} \right)^{T}\dot{p}(t),\notag\\ 
    \dot{p}(t) &= - \Omega(t)^{\wedge}p(t) , \notag\\
    \Omega (t) &= - (J+J_{\phi})^{-1}J_{\phi}\dot{\phi}(t) + \left( J+J_{\phi} \right)^{-1} p(t),
\end{align*}
where the fact that $A$ and $I$ are constant was used and $f_{u} = \mathbb{M}^{-1}(\tau_{M})$, with $\tau^{a}_{M}= u^{a}(t)\mathrm{d}\phi_{a}\in \Gamma^{\infty}(T^{*}\mathbb{T}^{3})$, for $a=1,2,3$, is the torque applied to the reaction wheels.

Let $R_{d}\vcentcolon \mathbf{I}\to SO(3)$ be a desired curve, with desired body velocity $\Omega_{d}(t)\triangleq R^{T}_{d}(t)\dot{R}_{d}(t) \in \mathfrak{so}(3)$ for $t\in\mathbf{I}$, and $R_{e}(t) = R^{T}_{d}(t)R(t) \in SO(3)$ be the error velocity, which satisfies
\begin{align*}
    \dot{R}_{e}(t) &= R_{e}(t)\left(\Omega_{e}(t)\right)^{\wedge},\notag\\
    \Omega_{e}(t) &= \Omega (t) - R^{T}_{e}\Omega_{d}(t).
\end{align*}
A kinematic controller $(\Omega_{u} , V_{SO(3)})$ can be defined by the log operation as 
\begin{align*}
 \Omega_{u}(R_{e}) & = \mathrm{log}(R_{e})^{\vee},\\
 \mathrm{log}(R_{e}) &\triangleq  
 \left\{ 
 \begin{array}{ll}
    0_{3\times 3},  & R_{e} = I_{3},   \\
    \frac{\vartheta(R_{e})}{2\sin\left( \vartheta(R_{e})\right)} \left( R_{e} - R^{T}_{e}\right) ,  & R_{e} \neq I_{3},
 \end{array} \right. 
\end{align*}
with the Morse function $V_{SO(3)}(R_{e}) = 2-\sqrt{1+\mathrm{tr}(R_{e}(t))}$ \citep{2012Exponential}, where  $\vartheta(R_{e}) = \arccos{\left( \frac{1}{2}\left( \mathrm{tr}(R_{e})-1\right)\right)} \in (-\pi,\pi)$. It can be verified that this kinematic controller fulfills all the properties in Definition \ref{def:KinCtrl} 
with $\mathcal{O}_{R}\triangleq \{ R\in SO(3) \; | \; \mathrm{tr}(R)$ $=-1 \}$ and $\mathcal{U}_{R} \triangleq \{ R_{e}\in SO(3)\backslash$ $\mathcal{O}_{R} \;|\; V_{SO(3)}(R_{e})< 2-\epsilon \}$, for some $\epsilon>0$ arbitrarily small.

Then, by \eqref{eq:w} and  \eqref{eq:TckCtrl} the vector field $w\in T\mathbb{T}^{3}$ and the tracking controller can be calculated as
\begin{align*}
    w(t) =&  - J^{-1}_{\phi}(J+J_{\phi}) \big( \lambda \Omega_{u}(R_{e}) - R^{T}_{e}\Omega_{d} \\
&+ (J+J_{\phi})^{-1}p(t) \big),\\
    f_{u} =& -\dot{w}(t) - \frac{1}{2}\mathbb{M}^{-1}(h(\cdot,\cdot)) - k_{s}\varsigma (t) \\
    &+ \mathbb{M}^{-1}\left( ((J+J_{\phi})^{-1}J_{\phi})^{T} \psi (R_{e})\left(R_{e}-R^{T}_{e}\right)^{\vee}  \right) ,\notag\\
    \varsigma (t)&= \dot{\phi}(t)+w(t), \ \ 
    \psi(R_{e}) = \frac{1}{2\sqrt{1+\mathrm{tr}(R_{e})}},
\end{align*}
for some positive constants $\lambda$ and $k_{s}$, where the fact that $\mathbb{M}$, $A$, and $I$ do not depend on $\phi(t)$ was used. In addition, the torque for each wheel is given by $\tau_{M} = \mathbb{M}f_{u}$. Thus, in view of Theorem \ref{thm:TckCtrl}, this controller ensures the exponential convergence $(R(t), \Omega(t))\to (R_{d}(t),\Omega_{d}(t))$ for all $(R(0),\Omega(0))\in \mathcal{U}_{R}\times \mathfrak{so}(3)$, and the asymptotic convergence for all $(R(0),\Omega(0))\in SO(3)\backslash\mathcal{O}_{R}\times \mathfrak{so}(3)$.

\subsection{Unicycle mobile robot}
\begin{figure}
\begin{center}
\includegraphics[scale=1,trim = 0mm 0mm 0mm 0mm]{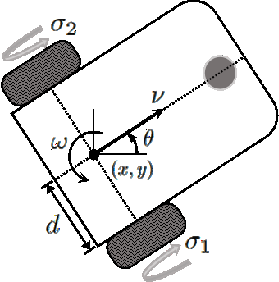}
\caption{Two-wheeled mobile robot scheme.}  
\label{fig:MobileR}                                 
\end{center}                                 
\end{figure}

The pose of a unicycle mobile robot, shown in Fig. \ref{fig:MobileR}, is given by 
$g=(\mathbf{p},R)\in SE(2)$, where $SE(2)=\{(\mathbf{p},R)\ |\ \mathbf{p}=[x \ y]^T\in \mathbb{R}^2, R\in SO(2)\}$ denotes the Lie group of rotations and translations in the plane, where 
\begin{equation*}
    R(\theta) = \begin{bmatrix}
         \cos{(\theta)}& -\sin{(\theta)}  \\
     \sin{(\theta)} & \cos{(\theta)} 
    \end{bmatrix} .  
\end{equation*}

The group operation for $SE(2)$ is $g_{1}g_{2}=(R_{1}\mathbf{p}_{2}+\mathbf{p}_{1},R_{1}R_{2})\in SE(2)$ for $g_{1}=(\mathbf{p}_{1},R_{1})$ and $g_{2}=({p}_{2},R_{2})\in SE(2)$. The identity element is $(0_{2\times 1},I_{2})\in SE(2)$, and the inverse for $g=(\mathbf{p},R)\in SE(2)$ is $g^{-1}=(-R^{T}\mathbf{p}, R^{T})$.

The two wheels are independently driven and have the coordinate $\sigma = (\sigma_{1},\sigma_{2})\in\mathbb{T}^{2}$. The robot configuration manifold is therefore $Q = SE(2)\times \mathbb{T}^{2}$. 
An action group $\Phi (g_{a},q) = (g_{a}g , \sigma ) \in Q$, for $q=(g,\sigma)\in Q$, and some $g_{a}\in SE(2)$ can be defined, resulting in  
$(Q , SE(2), \Phi , \mathbb{T}^{2})$ as a principal fiber bundle.

The Lie algebra of $SE(2)$ is expressed as
\begin{align*}
    \mathfrak{se}(2) &= \left\{ \left[ \begin{array}{cc}
        \omega^{\wedge} & \mathbf{v}  \\
         0_{1\times 2} & 0
    \end{array} \right] \in\mathbb{R}^{3\times 3} \; | \; \omega^{\wedge}\in \mathfrak{so}(2) \; , \mathbf{v}\in\mathbb{R}^{2} \right\} , \notag \\ 
    \omega ^{\wedge} &= \left[ \begin{array}{cc}
         0 & -\omega  \\
         \omega & 0 
    \end{array} \right] , \quad \forall\omega \in\mathbb{R},
\end{align*}
which is isomorphic to $\mathbb{R}^{3}$, $\mathfrak{se}(2)\simeq\mathbb{R}^{3}$. Let $(\cdot)^{\veebar}\vcentcolon \mathfrak{se}(2)\to \mathbb{R}^{3}$, and $(\cdot)^{\barwedge} \vcentcolon \mathbb{R}^{3}\to\mathfrak{se}(2)$, then  the adjoint operator satisfies $\mathrm{ad}_{\xi^{\barwedge}_{1}}\xi^{\barwedge}_{2} = ( \omega^{\wedge}_{1}\mathbf{v}_{2}-\omega_{2}^{\wedge}\mathbf{v}_{1} , 0 )^{\barwedge} \in\mathfrak{se}(2)$, for $\xi_{1} = (\mathbf{v}_{1}, \omega_{1})\in \mathbb{R}^{3}$, $\xi_{2} = (\mathbf{v}_{2}, \omega_{2})\in \mathbb{R}^{3}$. For each curve $g\vcentcolon \mathbf{I}\to SE(2)$, the body velocity without constraints is defined by
\begin{align*}
    \xi (t) &\triangleq \left(g^{-1}(t)\dot{g}(t)\right)^{\veebar} = ( \mathbf{v}(t) , \omega (t) ) , \notag\\
    \mathbf{v}(t) &=  R^{T}(t)\dot{\mathbf{p}}(t), \; \omega (t) = \left(R^{T}(t)\dot{R}(t)\right)^{\vee} = \dot{\theta}(t) ,
\end{align*}
for all $t\in \mathbf{I}$, where $(\cdot)^{\vee} \vcentcolon \mathfrak{so}(2)\to \mathbb{R}$. 

The no-slip condition leads to the following constraints:
\begin{align*}
    \dot{x}(t)\cos{(\theta )} + \dot{y}(t) \sin{(\theta )}  - \frac{\rho}{2}\left( \dot{\sigma}_{1}(t) + \dot{\sigma}_{2}(t)  \right) &= 0 ,\notag\\
    -\dot{x}(t)\sin{(\theta )} + \dot{y}(t)\cos{(\theta )} &= 0 ,\\ 
    \dot{\theta}(t) - \frac{\rho}{2d}\left( \dot{\sigma}_{1}(t) - \dot{\sigma}_{2}(t)  \right) &=0 ,
\end{align*}
 where $\rho >0$ is the radius of the wheels. Notice that the constraint can be rearranged as $\mathrm{w}(q) \dot{q} = 0_{5\times 1}$ for all 
 $q=(g,\sigma)\in Q$ and is $SE(2)$-invariant, i.e., $\mathrm{w}(\Phi_{g_{a}}q)T_{q}\Phi_{g_{a}}\dot{q} = \mathrm{w}(q) \dot{q} = 0_{5\times 1}$, for any $g_{a}\in SE(2)$.

 Thus, body velocities under the constraint are elements of the set $\overline{\mathfrak{se}}(2) \triangleq \{ \xi^{\barwedge}=(\mathbf{v},\omega)^{\barwedge}\; |\: \mathbf{v}=[\nu, 0]^{T} \in\mathbb{R}^{2} \} \subset \mathfrak{se}(2)$. Therefore, the velocity of the mobile robot in the body frame can be expressed as $\xi (t) = (\nu (t),0,\omega (t))$, with $\nu (t)\in\mathbb{R}$ being the velocity in the direction of the $x$ axis of the body frame (cf. Fig. \ref{fig:MobileR}). Moreover, since the three equations of the constraint are independent and invariant, it is sufficient to define a nonholonomic mechanical connection $\mathbb{A}\vcentcolon T\mathbb{T}^{2}\to\overline{\mathfrak{se}}(2)$, such that $\xi^\barwedge(t) = -\mathbb{A}\dot{\sigma}(t)$. In matrix form $\mathbb{A}$ and its general inverse $\mathbb{A}^{\dagger}\vcentcolon \overline{\mathfrak{se}}(2)\to T\mathbb{T}^{2}$ are  
 \begin{align*}
    [\mathbb{A}] &= \left[ \begin{array}{cc}
        -\frac{\rho}{2}  & -\frac{\rho}{2}  \\
         0 & 0 \\
         -\frac{\rho}{2d} & \frac{\rho}{2d}
     \end{array}\right] , \quad \left[\mathbb{A}^{\dagger}\right] = \left[ \begin{array}{ccc}
        -\frac{1}{\rho} & 0   & -\frac{d}{\rho}  \\
         -\frac{1}{\rho} & 0 & \frac{d}{\rho}
     \end{array}\right] .
 \end{align*}
 
Let $\mu >0$, $J_{R}>0$, and $J_{\sigma}>0$ be the mass of the robot, the inertia of the robot about its center of mass, and the inertia of the wheels about the pivot point, respectively. Then, a metric on $Q$ is given by
\begin{align*}
\mathbb{G} &= \mu \left(\mathrm{d}x \otimes \mathrm{d}x + \mathrm{d}y \otimes \mathrm{d}y \right) + J_{R} (\mathrm{d}\theta \otimes \mathrm{d}\theta ) \notag\\
&\quad + J_{\sigma} \left(\mathrm{d}\sigma_{1} \otimes \mathrm{d}\sigma_{1} + \mathrm{d}\sigma_{2} \otimes \mathrm{d}\sigma_{2} \right) ,
\end{align*}
which is $SE(2)$-invariant. Therefore, the Lagrangian of the system is given by the kinetic energy associated with this metric, i.e.,  
\begin{equation*}
    L(q,\dot{q}) = \frac{1}{2}\mu \left( \dot{x}^{2}+\dot{y}^{2} \right) + \frac{1}{2}J_{R}\dot{\theta}(t) + \frac{1}{2}J_{\sigma}\left( \dot{\sigma}^{2}_{1}+\dot{\sigma}^{2}_{2} \right),
\end{equation*}
leading to the reduction $L(\Phi_{g^{-1}}q, T_{q}\Phi_{g^{-1}}\dot{q}) = l(\xi , \sigma, \dot{\sigma})$: 
\begin{align*}
    l(\xi , \sigma, \dot{\sigma}) &= \frac{1}{2}\left[ \xi^{T} , \dot{\sigma}^{T}  \right] \left[ \begin{array}{cc}
         I & 0_{3\times 2}  \\
         0_{2\times 3} & m
    \end{array} \right] \left[ \begin{array}{c}
         \xi\\
         \dot{\sigma}
    \end{array} \right] , \notag\\
    [I] &= \left[ \begin{array}{cc}
         \mu I_{2} & 0_{2\times 1}  \\
         0_{1\times 2} & J_{R}
    \end{array} \right] ,\quad [m] = J_{\sigma}I_{2}.
\end{align*}

In view of \eqref{eq:lag}, it follows that $A = 0$, and consequently $\overline{\mathbb{M}} = m + \mathbb{A}^{T}I\mathbb{A}$. Therefore, given a curve $\sigma \vcentcolon \mathbf{I}\to T\mathbb{T}^{2}$, the reduced equations of motion for the mobile robot are calculated by \eqref{eq:CnShapeDyn} and \eqref{eq:Cxi} as follows.
\begin{align}
    \ddot{\sigma} (t) &= \bar{f}_{u}, \label{eq:DynRM}\\
   \xi (t) &= -[\mathbb{A}]\dot{\sigma}(t), \quad \forall t\in\mathbf{I} \label{eq:KinRM},
\end{align}
where the fact that $I$, $A$ and $\overline{{\mathbb{M}}}$ are constant was used, and $\bar{f}_{u} = \overline{\mathbb{M}}^{-1}(\tau_{M}) \in \Gamma^{\infty}(T\mathbb{T}^{2})$. In addition, it is assumed that the control torques are applied to the wheels through $\tau^{a}_{M} = u^{a}(t)\mathrm{d}\sigma_{a}$, for $a=1,2$.

Let $g_{d}\vcentcolon \mathbf{I}\to SE(2)$ be a desired trajectory with body velocity $g_{d}(t) = (\mathbf{p}_{d}(t),R_{d}(t))\in SE(2)$ and $\xi_{d} (t) = \left(g^{-1}_{d}(t)\dot{g}_{d}(t)\right) ^{\veebar} = (\nu_{d}(t),0,\omega_{d}(t))$, which is assumed to be bounded, 
and $g_{e} = g^{-1}_{d}g \in SE(2)$ be the left error. Then the error $g_{e} = (\mathbf{p}_{e},R_{e}) = (R^{T}_{d}(\mathbf{p} - \mathbf{p}_{d}) , R^{T}_{d}R)$ satisfies
\begin{align*}
    \xi_{e}(t) &= \left(g^{-1}_{e} \dot{g}_{e}\right)^{\veebar} = \xi(t) - \left(\mathrm{Ad}_{g^{-1}_{e}}\xi^{\barwedge}_{d}\right)^{\veebar} =  (\mathbf{v}_{e},\omega_{e}), \notag\\
    \mathbf{v}_{e} (t) &= \mathbf{v}(t) - R^{T}_{e} \left( \mathbf{v}_{d} (t) + \omega_{d}^{\wedge}\mathbf{p}_{e}(t)  \right) ,\notag\\
    \omega_{e} (t) &= \omega(t) -\omega_{d}(t) ,
\end{align*}
where $\mathbf{v}(t)=[\nu (t) , 0]^{T}$ and $\mathbf{v}_{d}(t)=[\nu_{d}(t) , 0]^{T}$. Motivated by \cite{tayefi2019logarithmic}, the kinematic controller $(\xi_{u},V_{SE(2)})$ is proposed with
\begin{align}
    \xi_{u} (g_{e}) &= \big( z_{x}, 0 ,
     \theta_{e} + k_{b}\beta(\mathbf{v}_{z}) \big) \in\mathbb{R}^{3} ,\label{eq:CtrlKin1RM}\\
    V_{SE(2)}(g_{e}) &= \frac{k_{1}}{2}\|z\|^{2} + \frac{k_{2}}{2}\beta^{2}(\mathbf{v}_{z}), \label{eq:CtrlKin2RM}
\end{align}
for some positive constants $k_{b}$, $k_1$, and $k_{2}$, where $\theta_{e}(t) = \theta(t) - \theta_{d}(t)$, $\beta (\mathbf{v}_{z}) = -\arctan (z_{y}/z_{x})$,  $z=(\mathbf{v}_{z},\theta_{e})\triangleq \left(\mathrm{log}(g_{e})\right)^{\veebar}$,   $\mathbf{v}_{z} = [z_{x},z_{y}]^{T}\in\mathbb{R}^{2}$, and  $\mathrm{log}\vcentcolon SE(2)\to \mathfrak{se}(2)$, which  is the logarithm map on $SE(2)$ defined as
\begin{align*}
    \mathrm{log}(g) &\triangleq  \left[ \begin{array}{ccc}
         \theta^{\wedge} & E^{-1}(\theta)\mathbf{p} \\
         0_{1\times 2} & 0
    \end{array}  \right] \in\mathfrak{se}(2),\notag\\
    E^{-1}(\theta) &= \left[ \begin{array}{cc}
         \alpha (\theta)&  \frac{\theta}{2} \\
         -\frac{\theta}{2}& \alpha (\theta)
    \end{array}\right] , \quad \alpha (\theta) = \frac{\theta}{2}\cot{\left( \frac{\theta}{2}\right)}.
\end{align*}
for all $g=(\mathbf{p},R(\theta))\in SE(2)$,  provided that $\mathrm{tr}(g)\neq -1$. 

Note that $(\mathrm{log}(e))^{\veebar} = (0,0,0)$ at the identity $e = (0_{1\times 2},R(0))=(0_{1\times 2},I_{2}) $. Moreover, the kinematic controller $(\xi_{u},V_{SE(2)})$ fulfills the properties in Definition \ref{def:KinCtrl} and the constraints 
as shown in Appendix \ref{App:nhKinCtrl}. In fact, it guarantees the exponential convergence of $g_{e}(t)\to e$, for all $g_{e}(0)\in SE(2)\backslash \mathcal{O}_{SE(2)}$, where $\mathcal{O}_{SE(2)} \triangleq \{ g_{e}\in SE(2) \;|\; \theta_{e}= \pm \pi, \; \mathbf{p}_{e}= [ \mathbf{p}_{e,x} , 0]^{T} \}$, by choosing $k_{b}>2$, and $k_1,k_2>0$.
Likewise, the Morse function \eqref{eq:CtrlKin2RM} satisfies condition \eqref{eq:nhCondition} with constant $\gamma_{SE(2)}= \frac{k_{1}}{8}$, for all $g_{e}\in SE(2)\backslash \mathcal{O}_{SE(2)}$, as shown in Appendix \ref{App:nhCondition}. 

Therefore, let the vector field $\overline{w}(t)\in \Gamma^{\infty}\left(T\mathbb{T}^{2}\right)$ on the shape space be defined as
\begin{equation}\label{eq:wRM}
    \overline{w}(t) \triangleq -\left[\mathbb{A}^{\dagger}\right]\left( \lambda\xi_{u}(g_{e}) -\xi_{d} (t)\right),
\end{equation}
then, by \eqref{eq:nhTckCtrl} the nonholonomic controller is 
\begin{align}
    \bar{f}_{u} =& -\dot{\overline{w}}(t) - k_{s}\bar{\varsigma}(t)
    +\left.\overline{\mathbb{M}}^{-1}\right( \mathbb{A}^{T} \big( k_{1}H^{T}(z)z  \notag \\
    & \left.- k_{2} \overline{H}^{T}(z)\left( \beta/\|\mathbf{v}_{z}\|^{2}\right)^{\wedge} \mathbf{v}_{z}\big) \right) ,\label{eq:CtrlRM}\\
    \bar{\varsigma}(t)&= \dot{\sigma}(t) + \overline{w}(t) , \label{eq:VarSigRM}
\end{align}
for some positive constants $k_{s}$ and $\lambda$. Therefore, by Theorem \ref{thm:nhTckCtrl} the exponential convergence $(g(t),\xi(t))\to (g_{d}(t),\xi_{d}(t))$ is guaranteed for all $(g_{e}(0),\xi_{e}(0))\in SE(2)\backslash\mathcal{O}_{SE(2)}\times \mathbb{R}^{3}$.

The proposed sliding mode control \eqref{eq:CtrlRM} in closed loop with the unicycle mobile robot \eqref{eq:DynRM}-\eqref{eq:KinRM} was simulated to follow a desired trajectory $\xi_{d}(t)$, given by a Lemniscate-type curve calculated as $x_{d}(t) = 0.8 \cos{(0.1t)}$ (m), $y_{d}(t) = 0.6\sin{(0.2t)}$ (m). 
For such a purpose, the controller \eqref{eq:CtrlRM} 
is expressed in a covector field form 
\begin{align}
    \tau_{1} &= -\overline{\mathbb{M}}\left(\dot{\overline{w}}(t) +k_{s}\bar{\varsigma}\right) \notag\\
    & [\mathbb{A}]^{T}\left( k_{1}H^{T}(z)z - k_{2} \overline{H}^{T}(z)\left( \beta/\|\mathbf{v}_{z}\|^{2}\right)^{\wedge} \mathbf{v}_{z}\right),\label{eq:CtrlET}
\end{align}
where $\bar{\varsigma}$, $\overline{w}$, and kinematic controller $(\bar{\xi}_{u} , V_{SE(2)})$ are defined in \eqref{eq:VarSigRM}, \eqref{eq:wRM} and \eqref{eq:CtrlKin1RM}-\eqref{eq:CtrlKin2RM}, respectively. 
The parameters used in the simulation for the sliding mode controller \eqref{eq:CtrlET} were 
 $\lambda=1.5$, 
 $k_{s}=2.2$, 
 $k_{b}=10$ 
 $k_{1}=0.01$ 
 $k_{2}=0.1$.
The parameters of the robot and initial conditions are given in Table \ref{tab:Parameters}.    
\begin{table}[h]
 \caption{Parameters and initial conditions. \label{tab:Parameters}}
 \centering
 \begin{tabular}{@{}lll@{}}
 \hline
 Initial condition & Value & Unit\\
 \hline
 $\sigma(0)$ & $0_{2\times 1}$ & rad/s\\
 $\dot{\sigma}(0)$ & $0_{2\times 1}$ & rad/$\mathrm{s}^{2}$\\
 $\theta(0)$ & $-\pi/4$ & rad \\
 $\mathbf{p}(0)$ & $[-1,-1]^{T}$ & m\\
 \hline
 Robot parameters & Value & Unit \\\hline
 $J_{R}$ & $0.025$& Kgm$^{2}$\\
 $m$ & $3.0$ & Kg\\
 $J_{\sigma}$ & $6\times 10^{-5}$ & Kgm$^{2}$ \\
 $\rho$ & $0.05$ & m \\
 $d$ & $0.165$ & m \\
 \hline
 \end{tabular}
 \end{table}

\begin{figure}
\begin{center}
\includegraphics[scale=0.15,trim = 40mm 3mm 20mm 0mm]{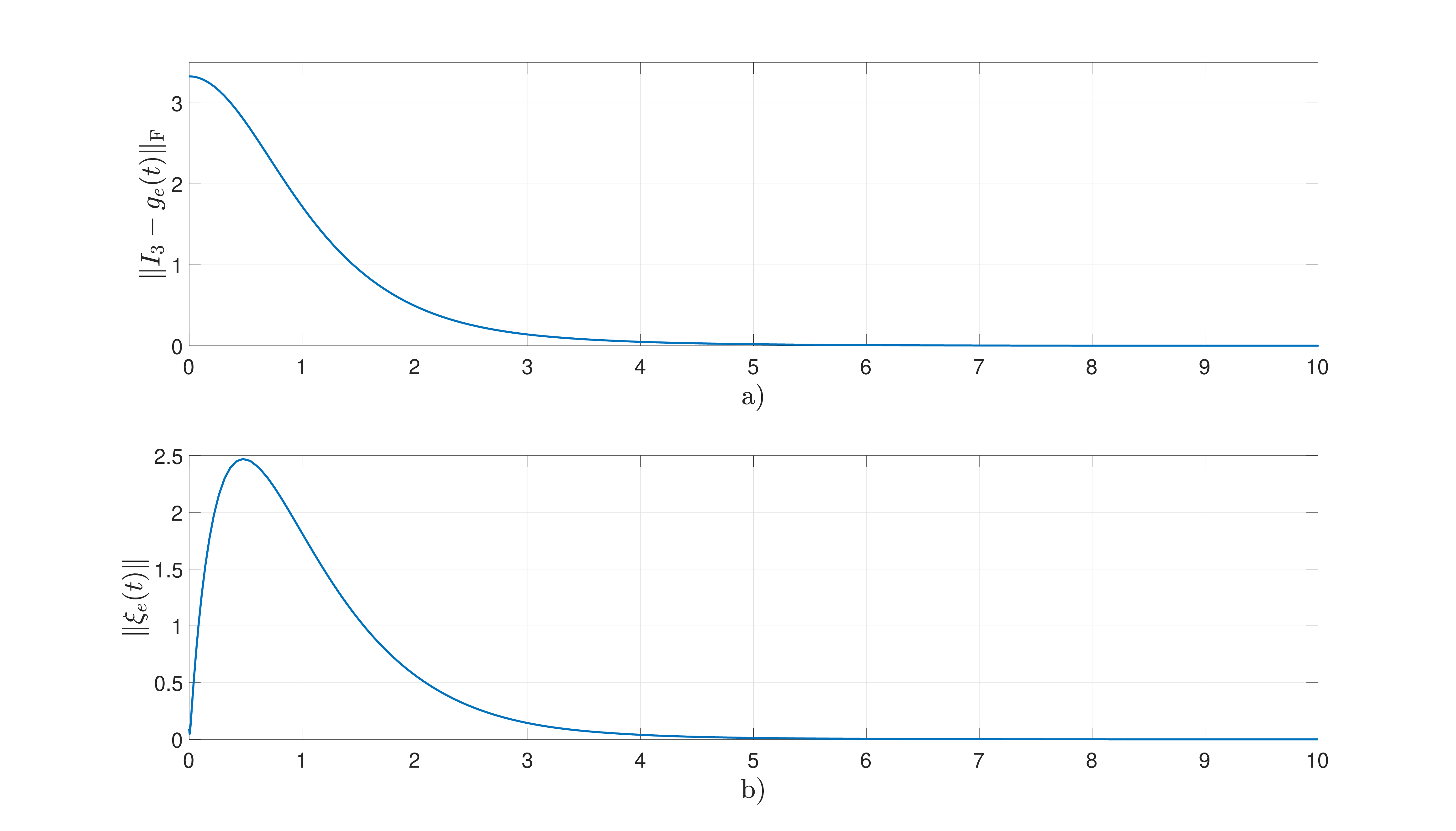}
\includegraphics[scale=0.15,trim = 40mm 0mm 20mm 3mm]{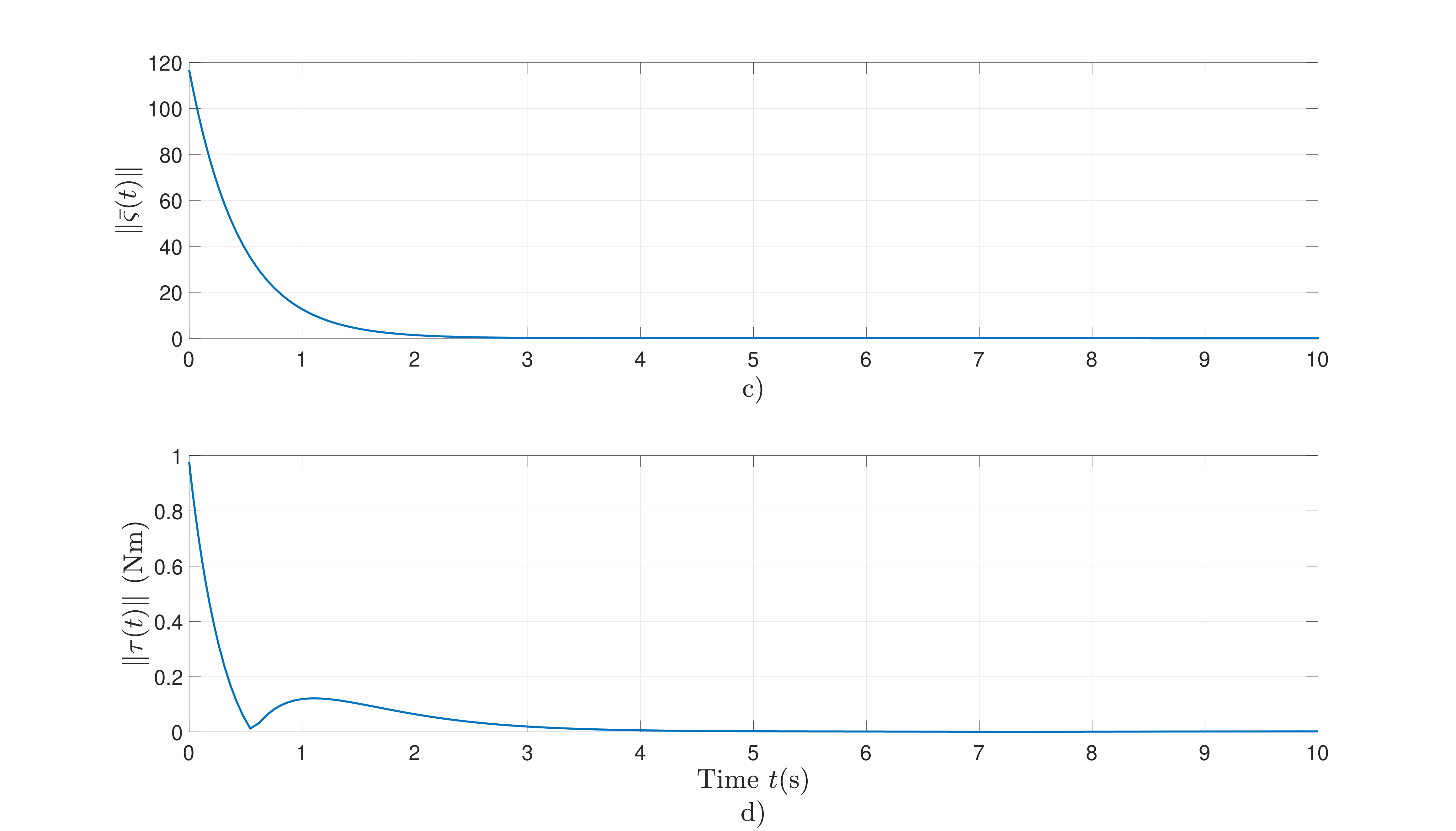}
\caption{Tracking performance of the proposed controller.}  
\label{fig:Sim1}                                 
\end{center}                                 
\end{figure}

Figure \ref{fig:Sim1} shows the behavior of the controller in tracking the desired trajectory. The position error calculated by the Frobenius norm $\|I_{3}-g_{e}\|_{\mathrm{F}}$ is shown in Fig. \ref{fig:Sim1}(a), which goes to zero exponentially. 
The norm of the velocity error $\|\xi_{e}(t)\|$ and the vector field $\|\bar{\varsigma}(t)\|$ in \eqref{eq:VarSigRM} are shown in Figs. \ref{fig:Sim1}(b)-(c). Finally, the control effort $\|\tau(t)\|$ is shown in Fig. \ref{fig:Sim1}(d). 

\section{Conclusions}\label{sec:Concl}

We have presented a framework for the systematic development of sliding-mode controllers for symmetric mechanical systems, both unconstrained and under nonholonomic constraints, that evolve on a configuration manifold with a principal fiber bundle structure. The symmetries enable us to work with the reduced motion equations and develop a reaching controller on the base manifold, thus simplifying the design and avoiding choosing a coordinate representation for the Lie (pose) group. 
Upon reaching the sliding subgroup, the sliding variable converges to the identity of the state manifold, achieving the control objectives. We applied the proposed design framework to two systems; numerical simulations were included for illustration.


\bibliographystyle{plainnat}  
\bibliography{PFB-arX}           

\appendix
\noindent
{\bf Appendix}

\section{Decomposition of Levi-Civita (LC) connection}\label{App:LCterms}    
The following lemma from \cite{cortes2002simple} is included here for a self-contained exposition. 

\begin{lem}
[Decomposition of LC connection] \label{lem:DecLC}
Let $(Q,G,\Phi, M)$ be a principal fiber bundle and $\mathbb{G}$, a Riemannian metric. The covariant derivative of $Y=(g\zeta , v)\in TQ$, with respect to $X=(g\xi , u)\in TQ$, for all $\xi(r),\zeta(r) \in\mathfrak{g}$ and $u,v\in TM$ , is given by
\begin{align}\label{eq:uConn}
    \overset{\mathbb{G}}{\nabla}_{X}Y &= \left( \begin{array}{c}
         \left( \overset{\mathbb{G}}{\nabla}_{X}Y \right)_{G}\\
         \left( \overset{\mathbb{G}}{\nabla}_{X}Y \right)_{M}
    \end{array} \right) \notag\\
    &=g\left( 
    \left(\begin{array}{c}
        \overset{I}{\nabla}_{\Omega}\Psi\\
        \overset{\mathbb{M}}{\nabla}_{u}v
    \end{array} \right) -\frac{1}{2} \left(\begin{array}{c}
        I^{\sharp}(\eta)\\
        \mathbb{M}^{\sharp}(h)
    \end{array} \right) 
    \right)  ,
\end{align}
where $\Omega = \xi + A(r)u$, $\Psi = \zeta + A(r)v$. The terms $\eta(X,Y) \in \mathfrak{g}^{*}$ and $h(X,Y)\in T^{*}M$ are given as follows.
\begin{align*}
    \eta &= -D(I\Omega)(\cdot,v) -D(I\Psi)(\cdot,u) + I\big( [\Omega, \Psi] - [\xi , \zeta] \\
    &\quad +\xi_{r}v - \zeta_{r}u - A[u,v]\big) + 2I\left( A  \left( \overset{\mathbb{G}}{\nabla}_{X}Y \right)_{M}\right) \in \mathfrak{g}^{*}, \\
    h&= I(\Omega , B(v,\cdot)) + I(\Psi , B(u,
    \cdot)) + DI(\cdot)(\Omega , \Psi) \in T^{*}M,
\end{align*}
with $\xi_{r} = \frac{\partial \xi}{\partial r}$ and $\zeta_{r} = \frac{\partial \zeta}{\partial r}$. The term $D$ denotes the derivative along connection $\mathcal{A}$ \citep{cortes2002simple}.

In particular, if $Y\in \mathcal{D}$ for some distribution $\mathcal{D}\subset TQ$, the nonholonomic affine connection is calculated as
\begin{equation}\label{eq:cConn}
    \overset{\mathbb{G}}{\overline{\nabla}}_{X}Y =g\left( 
    \left(\begin{array}{c}
        \mathcal{A}^{\mathrm{sym}}\left(\overset{I}{\nabla}_{\bar{\Omega}}\bar{\Psi}\right)\\
        \overset{\overline{\mathbb{M}}}{\nabla}_{u}v
    \end{array} \right) -\frac{1}{2} \left(\begin{array}{c}
        \bar{I}^{\sharp}(\bar{\eta}) \\
        \overline{\mathbb{M}}^{\sharp}(\bar{h})
    \end{array} \right) 
    \right)  ,
\end{equation} 
with $\bar{\Omega} = \xi + \mathbb{A}(r)u$, $\bar{\Psi} = \zeta + \mathbb{A}(r)v$, and  $\overline{\mathbb{M}}\triangleq \mathbb{M}+ \Tilde{A}^{T}I\Tilde{A}$, for $\Tilde{A} = A-\mathbb{A}$. 

Likewise, the terms $\bar{\eta}(X,Y) \in \overline{\mathfrak{g}}^{*}$ and $\bar{h}(X,Y)\in T^{*}M$ are calculated as 
\begin{align*}
    \bar{\eta} &= -\mathbb{D}(I\bar{\Omega})(\cdot,v) -\mathbb{D}(I\bar{\Psi})(\cdot,u) + I\big( [\bar{\Omega}, \bar{\Psi}] - [\xi , \zeta] \\
    &\quad +\xi_{r}v - \zeta_{r}u - \mathbb{A}[u,v]\big) + 2\overline{I}\left( \mathbb{A}  \left( \overset{\mathbb{G}}{\overline{\nabla}}_{X}Y \right)_{M}\right) \\
    &\quad +I\left( \Tilde{A}u , \lambda_{\cdot}v - [\cdot , \zeta] \right) + I\left( \Tilde{A}v , \lambda_{\cdot}u - [\cdot , \xi] \right), \\
    \bar{h}&= I(\bar{\Omega} , B(v,\cdot)) + I(\bar{\Psi} , B(u,
    \cdot)) + I(\Tilde{A}v,\mathbb{B}(u,\cdot)) \\
     &+ I(\Tilde{A}u,\mathbb{B}(v,\cdot)) -D(I\bar{\Psi})(\Tilde{A}\cdot , u) - D(I\bar{\Omega})(\Tilde{A}\cdot , v) \\
     &+ \mathbb{D}I(\cdot) (\bar{\Omega} + \Tilde{A}u , \bar{\Psi} + \Tilde{A}v) - \mathbb{D}I(\cdot)(\tilde{A}u,\tilde{A}v) \\
     &- I([\xi,\zeta], \tilde{A}\cdot) - I(\zeta_{r}u-\xi_{r}u , \tilde{A}\cdot) - I(\mathbb{A}[u,v], \tilde{A}\cdot) ,
\end{align*}
where the expression $\lambda_{\cdot}w$ for some $w\in T_{r}M$ is described by $\frac{\partial \underline{e}_{i}}{\partial r^{\alpha}} = \sum^{k}_{a=1}\lambda^{a}_{i\alpha}\underline{e}_{a}$, $\mathbb{D}$ is the derivative along the nonholonomic connection $\mathcal{A}^{\mathrm{nh}}$, and $\mathbb{B}$ is the local expression of the curvature for such connection \citep{cortes2002simple}.
\hfill$\blacksquare$
\end{lem}

\section{Nonholonomic kinematic controller}\label{App:nhKinCtrl}        
For $g_{e}=(\mathbf{p}_{e}, R(\theta_{e}))\in SE(2)$, Eq. \eqref{eq:CtrlKin1RM} can be rewritten as $\xi_{u}(g_{e})=(\Gamma \mathbf{v}_{z},\theta_{e}+k_{b}\beta(\mathbf{v}_{z}))$, where $\Gamma =\mathrm{diag} \{ 1,0\}\in\mathbb{R}^{2\times 2}$, $\mathbf{v}_{z}(g_{e}) =[z_{x},z_{y}]^{T}=E^{-1}(\theta_{e})\mathbf{p}_{e}$, and $\beta(\mathbf{v}_{z})=-\arctan(z_{y}/z_{x})$. Thus, at the identity $e=(0_{2\times 1}, R(0))$, it yields $\mathbf{v}_{z}(e) =E^{-1}(0)0_{2\times 1} = 0_{2\times 1} $ and consequently $\xi_{u}(e)= (0_{2\times 1}, 0)$ because $\beta(0,0)=-\arctan (0/0) = 0$. This shows that the kinematic controller has the property $(i)$ of Subsection \ref{sec:KinCtrl}. 

Likewise, for the inverse element $g^{-1}_{e}=\big(-R^{T}(\theta_{e})\mathbf{p}_{e},$ $ R(-\theta_{e})\big)\in SE(2)$, it has $ \mathbf{v}_{z}\big(g^{-1}_{e}\big) =  -E^{-1}(-\theta_{e})R^{T}(\theta_{e})\mathbf{p}_{e}$ $ = -\mathbf{v}_{z}(g_{e})$, 
where the fact that $E^{-1}(-\theta)R^{T}(\theta) =E^{-1}(\theta)$ was used. In addition, it can be verified that
\begin{align*}
\beta\left( \mathbf{v}_{z}\right) &=  \beta\left( g_{e}\right) = -\arctan\left( \frac{\sin{(\theta_{e})}}{1+\cos{(\theta_{e})}}\frac{\|\mathbf{p}\|+\|\mathbf{p}_{d}\|}{\|\mathbf{p}\|-\|\mathbf{p}_{d}\|} \right).
\end{align*}
Therefore, at $g^{-1}_{e}$ it fulfills 
\begin{align*}
\beta \left( g^{-1}_{e}\right) 
    &= -\arctan\left(\frac{\sin{(-\theta_{e})}}{1+\cos{(-\theta_{e})}}\frac{\|\mathbf{p}\|+\|\mathbf{p}_{d}\|}{\|\mathbf{p}\|-\|\mathbf{p}_{d}\|}  \right)\notag\\
    &=-\beta(g_{e}) = -\beta\left( \mathbf{v}_{z}\right),
\end{align*}
thus, $\xi_{u}\left(g^{-1}_{e}\right) = \left( \Gamma \mathbf{v}_{z}(g^{-1}_{e}),-\theta_{e}+k_{b}\beta(\mathbf{v}_{z}(g^{-1}_{e})  \right) = \left( -\Gamma \mathbf{v}_{z},-\theta_{e}-k_{b}\beta(\mathbf{v}_{z})  \right) = -\xi_{u}(g_{e})$, fulfilling the property $(ii)$ of kinematic controllers. 

Regarding property $(iii)$ and $(iv)$, consider the function \eqref{eq:CtrlKin2RM}, which satisfies
\begin{equation*}
\frac{k_{1}}{2}\|z\|^{2}    \leq V_{SE(2)}(g_{e}) \leq \lambda_{\max}(Q_{1})\|z\|^{2} , 
\end{equation*}
where 
\begin{equation*}
Q_{1}= \begin{bmatrix}
     \frac{1}{2} \left( k_{1}+k_{2}\frac{\beta^{2}}{\|\mathbf{v}_{z}\|^{2}}\right) I_{2}& 0_{2\times 1}  \\
     0_{1\times 2}& \frac{k_{1}}{2} 
\end{bmatrix}>0,
\end{equation*}
then
\begin{align*}
    &\left\langle \mathrm{d}V_{SE(2)}(g_{e}); -g_{e} \xi_{u}(g_{e}) \right\rangle \notag\\ 
    =&\left\langle\left\langle g^{-1}_{e} \mathrm{grad}V_{SE(2)}(g_{e}) , -\xi_{u}(g_{e}) \right\rangle\right\rangle \notag\\  
    =& -\left( k_{1}z^{T}H(z) + k_{2}\beta \mathbf{v}^{T}_{z}\left( 1/\|\mathbf{v}_{z}\|^{2} \right)^{\wedge}\overline{H}(z)\right)\xi_{u}(g_{e}),
\end{align*}
where the following time derivatives were used
\begin{align*}
    &\frac{\mathrm{d}}{\mathrm{d}t}z = \left[ \begin{array}{c}
         \dot{\mathbf{v}}_{z}\\
         \omega_{e} 
    \end{array} \right] = H(z)\xi_{e}, \notag\\
    &\frac{\mathrm{d}}{\mathrm{d}t}\beta (\mathbf{v}_{z}) = \mathbf{v}_{z}^{T}\left(\frac{1}{\|\mathbf{v}_{z}\|^{2}}\right)^{\wedge} \overline{H}(z)\xi_{e} , \notag\\
    &H(z) = \left[ \begin{array}{c}
         \overline{H}(z)\\
         \left[ 0, 0, 1 \right] 
    \end{array} \right] = \left[ \begin{array}{cc}
         E^{-T}(\theta_{e})  &\frac{1}{\theta_{e}}\left( I_{2} - E^{-T}(\theta_{e})\right) \mathbf{v}_{z}  \\
         0_{1\times 2}& 1 
    \end{array} \right].
\end{align*}
Therefore, substituting \eqref{eq:CtrlKin1RM} yields
\begin{align*}
    &\left\langle \mathrm{d}V_{SE(2)}(g_{e}); -g_{e} \xi_{u}(g_{e}) \right\rangle \notag\\
    =& - k_{1}\mathbf{v}^{T}_{z}Q_{2}\mathbf{v}_{z} - k_{1} \theta^{2}_{e} - k_{2}\mathbf{v}^{T}_{z}Q_{3}\mathbf{v}_{z},
\end{align*}
where
\begin{align*}
    Q_{2} &= \left[ \begin{array}{cc}
         \varphi_{1}(\theta_{e},\mathbf{v}_{z}) & \frac{\theta_{e}}{4}  \\
         \frac{\theta_{e}}{4} & \varphi_{2}(\theta_{e},\mathbf{v}_{z}) 
    \end{array} \right] , \notag\\
    Q_{3} &= \frac{1}{2}\left[ \begin{array}{cc}
         k_{b}\left(\beta^{2}/\|\mathbf{v}_{z}\|^{2}\right)& \left(\beta/\|\mathbf{v}_{z}\|^{2}\right) \alpha (\theta_{e})  \\
         \left(\beta/\|\mathbf{v}_{z}\|^{2}\right) \alpha (\theta_{e}) & \left(\beta/\|\mathbf{v}_{z}\|^{2}\right)  (\theta_{e}+k_{b}\beta) 
    \end{array} \right] ,
\end{align*}
with $\varphi_{1}(\theta_{e},\mathbf{v}_{z}) = 1+k_{b}\frac{\beta}{\theta_{e}}(1-\alpha (\theta_{e})) + k_{b} \left(\beta/\|\mathbf{v}_{z}\|^{2}\right)\theta_{e}$, and $\varphi_{2}(\theta_{e},\mathbf{v}_{z}) = (\theta_{e}+k_{b}\beta)\frac{1}{\theta_{e}}(1-\alpha (\theta_{e}))+ k_{b} \left(\beta/\|\mathbf{v}_{z}\|^{2}\right)\theta_{e}$, being $Q_{2}$ and $Q_{3}$ bounded matrices for each $\theta_{e}(t)\in (-\pi,\pi)$ and some $\beta\neq 0$. 
By a sufficiently large $k_{b}>2$, $Q_{3}$ is positive definite (\cite{tayefi2019logarithmic}). Then, it gives
\begin{align*}
    &\left\langle \mathrm{d}V_{SE(2)}(g_{e}); -g_{e} \xi_{u}(g_{e}) \right\rangle \notag\\
    \leq& -\gamma_{1}\|z\|^{2} \leq -\frac{\gamma_{1}}{\lambda_{\max}(Q_{1})}V_{SE(2)}(g_{e}),
\end{align*}
by defining $\gamma_{1} \triangleq \min \left\{ k_{2}\lambda_{\min}(Q_{3}) - k_{1}\lambda_{\max}(Q_{2}) , \; k_{1}\right\}>0$, for $k_2>k_1|\lambda_{max}(Q_2)|/\lambda_{min}(Q_3)$ and $k_1>0$ .  
Therefore, for all $z(g_{e}(0))=(\mathbf{v}_{z},\theta_{e})\in \mathcal{U}_{SE(2)}\triangleq \{ z(g_{e})\in \mathbb{R}^{3}\;|\; \theta_{e}\in (-\pi,\pi) , \; \beta(z)\neq 0 \}$, there exist constants $k_{b}>2$ and $k_{1,2}>0$ such that properties 
$(iii)-(iv)$ of kinematic controllers are fulfilled.

 \section{Nonholonomic Morse function condition}\label{App:nhCondition}
Given $g=(x,y, R(\theta)\in SE(2))$ and $\xi = (\nu_{x},\nu_{y},\omega)\in \mathfrak{se}(2)$, the adjoint map $\mathrm{Ad}\vcentcolon SE(2)\times \mathfrak{se}(2)\to \mathfrak{se}(2)$ can be described through the Lie algebra automorphism $\mathrm{Ad}_{g}\vcentcolon \mathfrak{se}(2)\to \mathfrak{se}(2)$ calculated as 
    \begin{equation*}
        \left( \mathrm{Ad}_{g}\xi \right)^{\veebar} = \left[ \begin{array}{ccc}
            \cos{\theta} & -\sin{\theta}  & y \\
             \sin{\theta}&\cos{\theta}  & -x \\
             0 & 0& 1
        \end{array} \right] \left[ \begin{array}{c}
             \nu_{x} \\
             \nu_{y}\\
             \omega
        \end{array} \right].
    \end{equation*}
Thus, the matrix form of the automorphism $\mathrm{Ad}_{e}, i.e., \mathrm{Ad}_{g_{e}}$,  is a $2$-rank matrix whose Frobenius norm satisfies
\begin{align*}
    &\|\mathrm{Ad}_{e}-\mathrm{Ad}_{g_{e}}\|^{2}\leq \|\mathrm{Ad}_{e}-\mathrm{Ad}_{g_{e}}\|^{2}_{F} \notag\\
    =&4\left( 1-\cos{\theta_{e}}\right)+\|\mathbf{p}_{e}\|^{2} \leq 4\left(\theta^{2}_{e}+\|\mathbf{p}_{e}\|^{2}\right).
\end{align*}
Therefore,  the Morse function \eqref{eq:CtrlKin2RM} fulfills Assumption \ref{Asm:nhKinCtrl}, i.e., 
\begin{align*}
    V_{SE(2)}(g_{e})\geq \frac{k_{1}}{2}\|z\|^{2}&\geq\frac{k_{1}}{2}\left(\|\mathbf{p}_{e}\|^{2} +\theta^{2}_{e}\right) \notag\\
    &\geq \frac{k_{1}}{8} \|\mathrm{Ad}_{e}-\mathrm{Ad}_{g_{e}}\|^{2},
\end{align*}
for all $g_{e}\in SE(2)\backslash\mathcal{O}_{SE(2)}$.
\end{document}